\documentclass[twoside,10pt]{article}

\usepackage{jmlr2e}

\jmlrheading{1}{2000}{1-48}{4/00}{10/00}{Chuanhao Li, Qingyun Wu and Hongning Wang}

\ShortHeadings{Unifying Clustered and Non-stationary Bandits}{Li, Wu and Wang}
\firstpageno{1}

\usepackage[utf8]{inputenc} 
\usepackage[T1]{fontenc}    
\usepackage{url}            
\usepackage{booktabs}       
\usepackage{amsfonts}       
\usepackage{nicefrac}       
\usepackage{microtype}      
\usepackage{graphicx}
\usepackage{algorithmic}
\usepackage{algorithm}
\usepackage{wrapfig}

\newcommand{\model}{{DyClu}}

\newtheorem{assumption}{Assumption}
\usepackage{amsmath}
\usepackage{amsfonts}
\DeclareMathOperator*{\argmax}{arg\,max}

\def \cU {\mathcal{U}}
\def \cN {\mathcal{N}}
\def \bU {\mathbb{U}}
\def \bO {\mathbb{O}}
\def \bM {\mathbb{M}}

\def \cH {\mathcal{H}}
\def \bA {\mathbf{A}}
\def \bx {\mathbf{x}}
\def \bX {\mathbf{X}}
\def \bY {\mathbf{Y}}
\def \by {\mathbf{y}}

\def \bb {\mathbf{b}}

\def \bI {\mathbf{I}}
\def \bR {\mathbb{R}}
\def \bE {\mathbb{E}}

\begin{document}

\title{Unifying Clustered and Non-stationary Bandits}

\author{\name Chuanhao Li \email cl5ev@virginia.edu \\
       \addr Department of Computer Science\\
       University of Virginia\\
       Charlottesville, VA 22903, USA\\
       \AND
       \name Qingyun Wu \email qw2ky@virginia.edu \\
       \addr Department of Computer Science\\
       University of Virginia\\
       Charlottesville, VA 22903, USA\\
       \AND
       \name Hongning Wang \email hw5x@virginia.edu \\
       \addr Department of Computer Science\\
       University of Virginia\\
       Charlottesville, VA 22903, USA
       }


\maketitle

\begin{abstract}
Non-stationary bandits and online clustering of bandits lift the restrictive assumptions in contextual bandits and provide solutions to many important real-world scenarios. Though the essence in solving these two problems overlaps considerably, they have been studied independently. In this paper, we connect these two strands of bandit research under the notion of test of homogeneity, which seamlessly addresses change detection for non-stationary bandit and cluster identification for online clustering of bandit in a unified solution framework. Rigorous regret analysis and extensive empirical evaluations demonstrate the value of our proposed solution, especially its flexibility in handling various environment assumptions. 
\end{abstract}


\section{Introduction}
\label{sec:intro}
Most existing contextual bandit algorithms impose strong assumptions on the mapping between context and reward \cite{abbasi2011improved,chu2011contextual, li2010contextual}: typically it is assumed that the expected reward associated with a particular action is determined by a \emph{time-invariant function} of the context vector. 
This overly simplified assumption restricts the application of contextual bandits in many important real-world scenarios, where a learner has to serve a population of users with possible mutual dependence and changing interest. This directly motivates recent efforts that postulate more specific reward  assumptions \cite{wu2016contextual,filippi2010parametric,maillard2014latent,kleinberg2008multi}; among them, non-stationary bandits \cite{wu2018learning,slivkins2008adapting,cao2019nearly,besson2019generalized,russac2019weighted,chen2019new} and online clustering of bandits \cite{gentile2014online,li2016collaborative,gentile2017context,li2019improved} received much attention.

In non-stationary bandits, the reward mapping function becomes time-variant.
A typical non-stationary setting is the abruptly changing environment, a.k.a, a piecewise stationary environment, in which the environment undergoes abrupt changes at unknown time points but remains stationary between two consecutive change points \cite{yu2009piecewise,garivier2011on}. A working solution needs to either properly discount historical observations \cite{hartland2006multi,garivier2011on,russac2019weighted} or detect the change points and reset the model estimation accordingly \cite{yu2009piecewise,cao2019nearly,wu2018learning}. In online clustering of bandits, 
grouping structures of bandit models are assumed in a population of users, e.g., users in a group share the same bandit model. But instead of assuming an explicit dependency structure, e.g., leveraging existing social network among users \cite{cesa2013gang,wu2016contextual}, online clustering of bandits aim to simultaneously cluster and estimate the bandit models during the sequential interactions with users \cite{gentile2014online,li2016collaborative,gentile2017context,li2019improved}. Its essence is thus to measure the relatedness between different users' bandit models. Typically, confidence bound of model parameter estimation \cite{gentile2014online} or reward estimation \cite{gentile2017context} is used for this purpose.   

So far these two problems have been studied in parallel; but the key principles to solve them overlap considerably. On the one hand, mainstream solutions for piecewise stationary bandits detect change points in the underlying reward distribution by comparing the observed rewards \cite{cao2019nearly} or the quality of estimated rewards \cite{yu2009piecewise,wu2018learning} in a window of consecutive observations. If change happens in the window, the designed statistics of interest would exceed a threshold with a high probability. This is essentially sequential hypothesis testing of a model's fitness \cite{siegmund2013sequential}. On the other hand, existing solutions for online clustering of bandits evaluate if two bandit models share the same set of parameters \cite{gentile2014online,li2016collaborative} or the same reward estimation on a particular arm \cite{gentile2017context}. This can also be understood as a goodness-of-fit test between models. 


In this work, we take the first step to unify these two parallel strands of bandit research under the notion of \emph{test of homogeneity}. We address both problems by testing whether the collection of observations in a bandit model follows the same distribution as that of new observations (i.e., change detection in non-stationary bandits) or of those in another bandit model (i.e., cluster identification in online clustering of bandits). Built upon our solution framework, bandit models can operate on individual users with much stronger flexibility, so that new bandit learning problems can be created and addressed. For example, learning in a clustered non-stationary environment, where the individual models are reset when a change of reward distribution is detected and merged when they are determined as identical. Our rigorous regret analysis and extensive empirical evaluations demonstrate the value of this unified solution, especially its advantages in handling various environment assumptions. 
\section{Related work}\label{sec:relatedWork}
Our work is closely related to the studies in non-stationary bandits and online clustering of bandits. In this section, we discuss the most representative solutions in each direction and highlight the connections between them.

\noindent\textbf{Non-stationary bandits.} 
Instead of assuming a time-invariant environment, the reward mapping is allowed to change over time in this problem setting. Commonly imposed assumptions include slowly-varying environment \cite{russac2019weighted, slivkins2008adapting} and abruptly-changing environment \cite{garivier2011on,wu2018learning}. We focus on the latter setting, which is also known as a piecewise stationary environment in literature \cite{yu2009piecewise,garivier2011on}. 
In a non-stationary setting, 
the main focus is to eliminate the distortion from out-dated observations, which follow a different reward distribution than that of the current environment. 
Mainstream solutions actively detect change points and reset bandit model accordingly \cite{yu2009piecewise, cao2019nearly, besson2019generalized,wu2018learning,hariri2015adapting}. 
As a typical example, \citet{wu2018learning} maintain a pool of base linear bandit models and adaptively add or select from them based on each model's reward estimation quality. This in essence boils down to a likelihood-ratio test for change detection. Other work directly uses cumulative sum control charts (CUSUM) \cite{hariri2015adapting} or generalized likelihood-ratio test \cite{besson2019generalized} for the purpose. In a nutshell, those employed tests essentially evaluate homogeneity of observation sequence over time.

\noindent\textbf{Online clustering of bandits.} 
Besides leveraging explicit structure among users, such as social networks \cite{buccapatnam2013multi,cesa2013gang,wu2016contextual}, recent efforts focus on online clustering of bandits via the interactions with users \cite{gentile2014online,li2016collaborative,gentile2017context,li2019improved}.  
For example, \citet{gentile2014online} assumed that observations from different users in the same cluster share the same underlying bandit parameter. Thus, they estimate the clustering structure among users based on the difference between their estimated bandit parameters. \citet{li2016collaborative} used a similar idea to cluster items (arms) as well. \citet{gentile2017context} further studied  arm-dependent clustering of users, by the projected difference between models on each arm. Essentially, these algorithms measure the relatedness between users by evaluating the homogeneity of observations associated with individual models, though they have used various measures for this purpose. 
\section{Methodology}
In this section, we first formulate the problem setup studied in this paper. Then we describe two key components pertaining to non-stationary bandits and online clustering of bandits, and pinpoint the essential equivalence between them under the notion of homogeneity test, which becomes the cornerstone of our unified solution.
Based on our construction of homogeneity test, we explain the proposed solution to the problem, followed by our theoretical analysis of the resulting upper regret bound of the proposed solution.

\subsection{Problem formulation}\label{subsec:problemFormulation}
To offer a unified approach that addresses the two target problems, we formulate a general bandit learning setting that encompasses both non-stationarity in individual users and existence of clustering structure among users.

Consider a learner that interacts with a set of $n$ users, $\cU=\left\{1,...,n\right\}$.
At each time $t=1,2,...,T$, the learner receives an arbitrary user indexed by $i_{t} \in \mathcal{U}$, together with a set of available arms $C_{t}=\{\bx_{t,1}, \bx_{t,2}, \dots, \bx_{t,K}\}$ to choose from, where $\bx_{t, j} \in \bR^d$ denotes the context vector associated with the arm indexed by $j$ at time $t$ (assume $\lVert \bx_{t,j}\rVert \leq 1$ without loss of generality), and $K$ denotes the size of arm pool $C_{t}$.
After the learner chooses an arm $\bx_{t}$, its reward $y_{t}\in \mathbb{R}$ is fed back from the user $i_{t}$.
We follow the linear reward assumption  \cite{abbasi2011improved,chu2011contextual, li2010contextual} and use $\theta_{i_{t},t}$ to denote the parameter of the reward mapping function in user $i_{t}$ at time $t$ (assume $\lVert \theta_{i_{t},t}\rVert \leq 1$). Under this assumption, reward at time $t$ is $y_{t} = \bx_{t}^\top \theta_{i_{t},t}+\eta_{t}$, where $\eta_{t}$ is Gaussian noise drawn from $N(0,\sigma^{2})$. Interaction between the learner and users repeats, and the learner's goal is to maximize the accumulated reward it receives from all users in $\cU$ up to time $T$.

Denote the set of time steps when user $i \in \cU$ is served up to time $T$ as $\cN_{i}(T)=\left\{1 \leq t \leq T:i_{t}=i\right\}$. Among time steps $t\in \cN_{i}(T)$, user $i$'s parameter $\theta_{i,t}$ changes abruptly at arbitrary time steps $\{c_{i,1},...,c_{i,\Gamma_{i}(T)-1}\}$, but remain constant between any two consecutive change points. $\Gamma_{i}(T)$ denotes the total number of stationary periods in $\cN_{i}(T)$. The set of unique parameters that $\theta_{i,t}$ takes for any user at any time is denoted as $\left\{\phi_{k}\right\}_{k=1}^{m}$ and their frequency of occurrences in $T$ is $\left\{p_{k}\right\}_{k=1}^{m}$. Note that the ground-truth linear parameters, the set of change points, the number and frequencies of unique parameters are unknown to the learner. Moreover, the number of users, i.e., $n$, and the number of unique bandit parameters across users, i.e., $m$, are finite but arbitrary.

Our problem setting defined above is general. The non-stationary and clustering structure of an environment can be specified by different associations between  $\{\theta_{i,t}\}^{n}_{i=1}$ and  $\left\{\phi_{k}\right\}_{k=1}^{m}$ across users over time $t=1,2,...,T$. For instance, by setting $n>m$ and $\Gamma_{i}(T)=1,\forall i \in \cU$, the problem naturally reduces to the online clustering of bandits problem, which assumes sharing of bandit models among users with stationary reward distributions. By setting $n=1$, $m>1$ and $\Gamma_{i}(T)>1,\forall i \in \cU$, it reduces to the piecewise stationary bandits problem, which only concerns a single user with non-stationary reward distributions in isolation.

\begin{figure}[t]
\vskip -1mm
\centerline{\includegraphics[width=\linewidth]{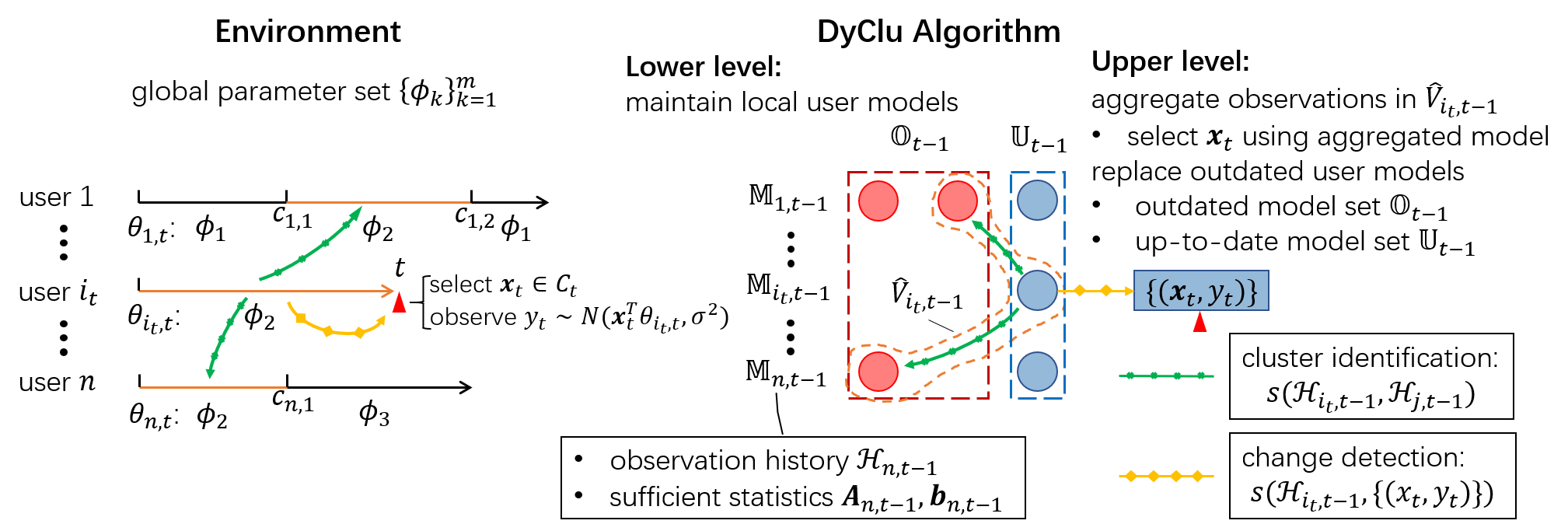}}
\vskip -1mm
\caption{Online bandit learning in a non-stationary and clustered environment. The environment setting is shown on the left side of the figure, where each user's reward mapping function undergoes a piecewise stationary process; and the reward mapping functions are globally shared across users. The proposed DyClu algorithm is illustrated on the right side of the figure. The model has a two-level hierarchy: at the lower level, individual users' bandit models are dynamically maintained; and at the upper level, a unified test of homogeneity is performed for the purpose of change detection and cluster identification among the lower-level user models.}
\label{fig:envAndAlgo}
\vskip -2mm
\end{figure}
To make our solution compatible with existing work in non-stationary bandits and clustered bandits, we also introduce the following three commonly made assumptions about the environment.
\begin{assumption}[Change detectability]\label{assump1}
For any user $i \in \cU$ and any change point $c$ in user $i$, there exists $\Delta>0$ such that at least $\rho$ portion of arms satisfy: $\lvert \bx^{\top}\theta_{i,c-1}-\bx^{\top}\theta_{i,c} \rvert > \Delta$ \citep{wu2018learning}.
\end{assumption}
\begin{assumption}[Separateness among $\left\{\phi_{k}\right\}_{k=1}^{m}$]\label{assump2}
For any two different unique parameters $\phi_{i} \neq \phi_{j}$, we have $\left\lVert \phi_{i}-\phi_{j}\right\rVert \geq \gamma >0$ \citep{gentile2014online,gentile2017context,li2019improved}.
\end{assumption}

\begin{assumption}[Context regularity]\label{assump3}
At each time $t$, arm set $C_{t}$ is generated i.i.d. from a sub-Gaussian random vector $X \in \bR^{d}$, such that $\mathbb{E}[XX^{\top}]$ is full-rank with minimum eigenvalue $\lambda'>0$; and the variance $\varsigma^{2}$ of the random vector satisfies $\varsigma^{2} \leq \frac{{\lambda'}^{2}}{8\ln{4K}}$ \citep{gentile2014online,gentile2017context,li2019improved}.
\end{assumption}

The first assumption establishes the detectability of change points in each individual user's bandit models over time. The second assumption ensures separation within the global unique parameter set shared by all users, and the third assumption specifies the property of context vectors.
Based on these assumptions, we establish the problem setup in this work and illustrate it on the left side of Figure \ref{fig:envAndAlgo}.

\subsection{Test statistic for homogeneity}\label{subsec:similarity_measure}
As discussed in Section \ref{sec:relatedWork}, the key problem in non-stationary bandits is to detect changes in the underlying reward distribution, and the key problem in online clustering of bandits is to measure the relatedness between different models. We view both problems as testing homogeneity between two sets of observations to unify these two seemingly distinct problems. For change detection, we test homogeneity between recent and past observations to evaluate whether there has been a change in the underlying bandit parameters for these two consecutive sets of observations. For cluster identification, we test homogeneity between observations of two different users to verify whether they share the same bandit parameters.
On top of the test results, we can operate the bandit models accordingly for model selection, model aggregation, arm selection, and model update. 

We use $\cH_{1}=\left\{(\bx_{i},y_{i})\right\}_{i=1}^{t_{1}}$ and $\cH_{2}=\left\{(\bx_{j},y_{j})\right\}_{j=1}^{t_{2}}$ to denote two sets of observations, where $t_{1},t_{2} \geq 1$ are their cardinalities. $(\bX_{1},\by_{1})$ and $(\bX_{2},\by_{2})$ denote design matrices and feedback vectors of $\cH_{1}$ and $\cH_{2}$ respectively, where each row of $\bX$ is the context vector of a selected arm and the corresponding element in $\by$ is the observed reward for this arm. Under our linear reward assumption, $\forall (\bx_{i},y_{i}) \in \cH_{1}$, $y_{i} \sim N(\bx_{i}^{\top}\theta_{1},\sigma^{2})$, and $\forall (\bx_{j},y_{j}) \in \cH_{2}$, $y_{j} \sim N(\bx_{j}^{\top}\theta_{2},\sigma^{2})$.
The test of homogeneity between $\cH_{1}$ and $\cH_{2}$ can thus be formally defined as testing whether $\theta_{1}=\theta_{2}$, i.e., whether observations in $\cH_{1}$ and $\cH_{2}$ come from a homogeneous population.

Because $\theta_{1}$ and $\theta_{2}$ are not observable, the test has to be performed on their estimates, for which maximum likelihood estimator (MLE) is a typical choice. Denote MLE for $\theta$ on a dataset $\cH$ as ${\vartheta}=(\bX^{\top}\bX)^{-}\bX^{\top}\by$, where $\left(\cdot\right)^{-}$ stands for generalized matrix inverse. A straightforward approach to test homogeneity between $\cH_{1}$ and $\cH_{2}$ is to compare $\lVert {\vartheta}_{1}-{\vartheta}_{2} \rVert$ against the estimation confidence on ${\vartheta}_{1}$ and ${\vartheta}_{2}$. 
The clustering methods used in \cite{gentile2014online,gentile2017context} essentially follow this idea. However, theoretical guarantee on the false negative probability of this method only exists when the minimum eigenvalues of $\bX_{1}^{\top}\bX_{1}$ and $\bX_{2}^{\top}\bX_{2}$ are larger than a predefined threshold. In other words, only when \emph{both} $\cH_{1}$ and $\cH_{2}$ have sufficient observations, this test is effective.

We appeal to an alternative test statistic that has been proved to be \emph{uniformly most powerful} for this type of problems \cite{chow1960tests,cantrell1991interpretation,wilson1978chow}:
\begin{equation}
\label{eq_chis_test}
s(\cH_{1},\cH_{2})=\frac{||\bX_{1}({\vartheta}_{1}-{\vartheta}_{1,2})||^{2}\!+\!||\bX_{2}({\vartheta}_{2}-{\vartheta}_{1,2})||^{2}}{\sigma^{2}}
\end{equation}
where ${\vartheta}_{1,2}$ denotes the estimator using data from both $\cH_{1}$ and $\cH_{2}$.
The knowledge about $\sigma^{2}$ can be relaxed by replacing it with empirical estimate, which leads to Chow test that has F-distribution \cite{chow1960tests}.

When $s(\cH_{1},\cH_{2})$ is above a chosen threshold $\upsilon$, it suggests the pooled estimator deviates considerably from the individual estimators on two datasets. Thus, we conclude $\theta_{1} \neq \theta_{2}$;
otherwise, we conclude $\cH_{1}$ and $\cH_{2}$ are homogeneous. The choice of $\upsilon$ is critical, as it determines the type-I and type-II error probabilities of the test. 
Upper bounds of these two error probabilities and their proofs are given below.

Note that $s(\cH_{1},\cH_{2})$ falls under the category of $\chi^{2}$ test of homogeneity. Specifically, it is used to test whether the parameters of linear regression models associated with two datasets are the same, assuming equal variance. It is known that this test statistic follows the noncentral $\chi^{2}$-distribution as shown in Theorem \ref{thm:similarityMeasureChiSquare} \citep{chow1960tests,cantrell1991interpretation}.

\begin{theorem}\label{thm:similarityMeasureChiSquare}
The test statistic $s(\cH_{1},\cH_{2})$ follows a non-central $\chi^{2}$ distribution $s(\cH_{1},\cH_{2}) \sim \chi^{2}(df, \psi)$
, where the degree of freedom $df=rank(\bX_{1})+rank(\bX_{2})-rank(\left[\begin{matrix}\bX_{1} \\ \bX_{2}\end{matrix}\right])$, and the non-centrality parameter $\psi=\frac{\left[\begin{matrix}\bX_{1}\theta_{1} \\ \bX_{2}\theta_{2}\end{matrix}\right]^{\top}\left[\bI_{t_{1}+t_{2}}-\left[\begin{matrix}\bX_{1} \\ \bX_{2}\end{matrix}\right]\left(\bX_{1}^{\top} \bX_{1}+\bX_{2}^{\top} \bX_{2}\right)^{-}\left[\begin{matrix}\bX_{1}^{\top} &\bX_{2}^{\top} \end{matrix}\right]\right]\left[\begin{matrix}\bX_{1}\theta_{1} \\ \bX_{2}\theta_{2}\end{matrix}\right]}{\sigma^{2}}$.
\end{theorem}

Then based on Theorem \ref{thm:similarityMeasureChiSquare}, the upper bounds for type I and type II error probabilities can be derived.

\begin{lemma}\label{lem:type1Similarity}
When $\theta_{1}=\theta_{2}$, $\psi=0$; the type-I error probability can be upper bounded by:
\begin{equation*}
P\big(s(\cH_{1},\cH_{2}) > \upsilon|\theta_{1}=\theta_{2}\big) \leq 1-F(\upsilon;df,0),
\end{equation*}
where $F(\upsilon;df,0)$ denotes the cumulative density function of distribution $\chi^{2}(df,0)$ evaluated at $\upsilon$.
\end{lemma}

\textit{Proof of Lemma \ref{lem:type1Similarity}.}

When datasets $\cH_{1}$ and $\cH_{2}$ are homogeneous, which means $\theta_{1}=\theta_{2}$, the non-centrality parameter becomes:
\begin{equation*}
\begin{split}
\psi & = \frac{1}{\sigma^{2}}\left[\begin{matrix}\bX_{1}\theta_{1} \\ \bX_{2}\theta_{1}\end{matrix}\right]^{\top}\left[\bI_{t_{1}+t_{2}}-\left[\begin{matrix}\bX_{1} \\ \bX_{2}\end{matrix}\right]\left(\bX_{1}^{\top} \bX_{1}+\bX_{2}^{\top} \bX_{2}\right)^{-}\left[\begin{matrix}\bX_{1}^{\top} &\bX_{2}^{\top} \end{matrix}\right]\right]\left[\begin{matrix}\bX_{1}\theta_{1} \\ \bX_{2}\theta_{1}\end{matrix}\right] \\
& = \frac{1}{\sigma^{2}}\left[\begin{matrix}\bX_{1}\theta_{1} \\ \bX_{2}\theta_{1}\end{matrix}\right]^{\top}\left[\begin{matrix}\bX_{1}\theta_{1} \\ \bX_{2}\theta_{1}\end{matrix}\right]-\frac{1}{\sigma^{2}}\left[\begin{matrix}\bX_{1}\theta_{1} \\ \bX_{2}\theta_{1}\end{matrix}\right]^{\top}\left[\begin{matrix}\bX_{1} \\ \bX_{2}\end{matrix}\right]\left(\bX_{1}^{\top} \bX_{1}+\bX_{2}^{\top} \bX_{2}\right)^{-}\left[\begin{matrix}\bX_{1}^{\top} &\bX_{2}^{\top} \end{matrix}\right]\left[\begin{matrix}\bX_{1}\theta_{1} \\ \bX_{2}\theta_{1}\end{matrix}\right] \\
& = \frac{1}{\sigma^{2}}\left[\theta_{1}^{\top}(\bX_{1}^{\top}\bX_{1}+\bX_{2}^{\top}\bX_{2})\theta_{1}-\theta_{1}^{\top}(\bX_{1}^{\top}\bX_{1}+\bX_{2}^{\top}\bX_{2})(\bX_{1}^{\top}\bX_{1}+\bX_{2}^{\top}\bX_{2})^{-}(\bX_{1}^{\top}\bX_{1}+\bX_{2}^{\top}\bX_{2})\theta_{1}\right] \\
& = \frac{1}{\sigma^{2}}\left[\theta_{1}^{\top}(\bX_{1}^{\top}\bX_{1}+\bX_{2}^{\top}\bX_{2})\theta_{1}-\theta_{1}^{\top}(\bX_{1}^{\top}\bX_{1}+\bX_{2}^{\top}\bX_{2})\theta_{1}\right]=0
\end{split}
\end{equation*}
Therefore, when $\theta_{1}=\theta_{2}$, the test statistic $s(\cH_{1},\cH_{2}) \sim \chi^{2}(df, 0)$. The type-I error probability can be upper bounded by $P(s(\cH_{1},\cH_{2}) > \upsilon|\theta_{1}=\theta_{2}) \leq 1-F(\upsilon;df,0)$, which concludes the proof of Lemma \ref{lem:type1Similarity}.

\begin{lemma}\label{lem:type2Similarity}
When $\theta_{1} \neq \theta_{2}$, $\psi \geq 0$; the type-II error probability can be upper bounded by,
\small
\begin{equation*}
    P\big(s(\cH_{1},\cH_{2}) \leq \upsilon|\theta_{1} \neq \theta_{2}\big) \leq \begin{cases}
             F\Big(\upsilon;d,\frac{||\theta_{1}-\theta_{2}||^{2}/\sigma^{2}}{{1}/{\lambda_{min}(\bX_{1}^{\top} \bX_{1})}+{1}/{\lambda_{min}(\bX_{2}^{\top} \bX_{2})}}\Big), & \text{if $\bX_{1}$ and $\bX_{2}$ are full-rank}.\\
            F(\upsilon;df,0), & \text{otherwise}.
          \end{cases}
\end{equation*}
\normalsize
\end{lemma}


\textit{Proof of Lemma \ref{lem:type2Similarity}.}

Similarly, using the cumulative density function of $\chi^{2}(df,\psi)$, we can show that the type-II error probability $P\big(s(\cH_{1},\cH_{2})\big) \leq \upsilon|\theta_{1} \neq \theta_{2}\big) \leq F(\upsilon;df,\psi)$. As mentioned in Section \ref{subsec:similarity_measure}, the value of $\psi$ depends on the unknown parameters $\theta_{1}$ and $\theta_{2}$. From the definition of $\psi$, we know that $\theta_{1}=\theta_{2}$ is only a sufficient condition for $\psi=0$. The necessary and sufficient condition for $\psi=0$ is that $\left[\begin{matrix}\bX_{1}\theta_{1} \\ \bX_{2}\theta_{2}\end{matrix}\right]$ is in the column space of $\left[\begin{matrix}\bX_{1} \\ \bX_{2}\end{matrix}\right]$, e.g., there exists $\theta$ such that $\left[\begin{matrix}\bX_{1}\theta_{1} \\ \bX_{2}\theta_{2}\end{matrix}\right]=\left[\begin{matrix}\bX_{1}\theta \\ \bX_{2}\theta\end{matrix}\right]$. 
Only when both $\bX_{1}$ and $\bX_{2}$ have a full column rank, $\theta_{1}=\theta_{2}$ becomes the necessary and sufficient condition for $\psi=0$. This means when either $\bX_{1}$ or $\bX_{2}$ is rank-deficient, there always exists $\theta_{1}$ and $\theta_{2}$, and $\theta_{1} \neq \theta_{2}$, that make $\psi=0$. For example, assuming $\bX_{1}$ is rank-sufficient and $\bX_{2}$ is rank-deficient, then $\psi=0$ as long as $\theta_{1}-\theta_{2}$ is in the null space of $\bX_{2}$.

To obtain a non-trivial upper bound of the type-II error probability, or equivalently a non-zero lower bound of the non-centrality parameter $\psi$, both $\bX_{1}$ and $\bX_{2}$ need to be rank-sufficient. Under this assumption, we can rewrite $\psi$ in the following way to derive its lower bound.

Denote $\epsilon=\theta_{2}-\theta_{1}$. Then $\theta_{2}=\theta_{1}+\epsilon$. We can decompose $\sigma^{2}\psi$ as:
\begin{equation*}
\begin{split}
\sigma^{2}\psi = & \left[\begin{matrix}\bX_{1}\theta_{1} \\ \bX_{2}(\theta_{1}+\epsilon)\end{matrix}\right]^\top\left[\bI_{t_{1}+t_{2}}-\left[\begin{matrix}\bX_{1} \\ \bX_{2}\end{matrix}\right]\left(\bX_{1}^\top\bX_{1}+\bX_{2}^\top\bX_{2}\right)^{-1}\left[\begin{matrix}\bX_{1}^\top & \bX_{2}^\top\end{matrix}\right]\right]\left[\begin{matrix}\bX_{1}\theta_{1} \\ \bX_{2}(\theta_{1}+\epsilon)\end{matrix}\right] \\
= & \left[\begin{matrix}\bX_{1}\theta_{1} \\ \bX_{2}\theta_{1}\end{matrix}\right]^\top\left[\bI_{t_{1}+t_{2}}-\left[\begin{matrix}\bX_{1} \\ \bX_{2}\end{matrix}\right]\left(\left[\begin{matrix}\bX_{1}^\top & \bX_{2}^\top\end{matrix}\right]\left[\begin{matrix}\bX_{1} \\ \bX_{2}\end{matrix}\right]\right)^{-1}\left[\begin{matrix}\bX_{1}^\top & \bX_{2}^\top\end{matrix}\right]\right]\left[\begin{matrix}\bX_{1}\theta_{1} \\ \bX_{2}\theta_{1}\end{matrix}\right] \\
& + \left[\begin{matrix}\bX_{1}\theta_{1} \\ \bX_{2}\theta_{1}\end{matrix}\right]^\top\left[\bI_{t_{1}+t_{2}}-\left[\begin{matrix}\bX_{1} \\ \bX_{2}\end{matrix}\right]\left(\left[\begin{matrix}\bX_{1}^\top & \bX_{2}^\top\end{matrix}\right]\left[\begin{matrix}\bX_{1} \\ \bX_{2}\end{matrix}\right]\right)^{-}\left[\begin{matrix}\bX_{1}^\top & \bX_{2}^\top\end{matrix}\right]\right]\left[\begin{matrix}0 \\ \bX_{2}\epsilon\end{matrix}\right] \\
& + \left[\begin{matrix}0 \\ \bX_{2}\epsilon\end{matrix}\right]^\top\left[\bI_{t_{1}+t_{2}}-\left[\begin{matrix}\bX_{1} \\ \bX_{2}\end{matrix}\right]\left(\left[\begin{matrix}\bX_{1}^\top & \bX_{2}^\top\end{matrix}\right]\left[\begin{matrix}\bX_{1} \\ \bX_{2}\end{matrix}\right]\right)^{-1}\left[\begin{matrix}\bX_{1}^\top & \bX_{2}^\top\end{matrix}\right]\right]\left[\begin{matrix}\bX_{1}\theta_{1} \\ \bX_{2}\theta_{1}\end{matrix}\right] \\
& + \left[\begin{matrix}0 \\ \bX_{2}\epsilon\end{matrix}\right]^\top\left[\bI_{t_{1}+t_{2}}-\left[\begin{matrix}\bX_{1} \\ \bX_{2}\end{matrix}\right]\left(\left[\begin{matrix}\bX_{1}^\top & \bX_{2}^\top\end{matrix}\right]\left[\begin{matrix}\bX_{1} \\ \bX_{2}\end{matrix}\right]\right)^{-1}\left[\begin{matrix}\bX_{1}^\top & \bX_{2}^\top\end{matrix}\right]\right]\left[\begin{matrix}0 \\ \bX_{2}\epsilon\end{matrix}\right] \\
\end{split}
\end{equation*}
Since $\left[\begin{matrix}\bX_{1}\theta_{1} \\ \bX_{2}\theta_{1}\end{matrix}\right]$ is in the column space of $\left[\begin{matrix}\bX_{1} \\ \bX_{2}\end{matrix}\right]$, the first term in the above result is zero. The second and third terms can be shown equal to zero as well using the property that matrix product is distributive w.r.t. matrix addition, which leaves us only the last term. Therefore, by substituting $\epsilon = \theta_{2}-\theta_{1}$ back, we obtain:
\begin{equation*}
\begin{split}
\psi 
& = \frac{1}{\sigma^{2}}(\theta_{1}-\theta_{2})^\top\bX_{2}^\top\bX_{2}(\bX_{1}^\top\bX_{1}+\bX_{2}^\top\bX_{2})^{-1}\bX_{1}^\top\bX_{1}(\theta_{1}-\theta_{2})\\
& \geq \frac{1}{\sigma^{2}}||\theta_{1}-\theta_{2}||^{2}\lambda_{\min}\Big(\bX_{2}^\top\bX_{2}(\bX_{1}^\top\bX_{1}+\bX_{2}^\top\bX_{2})^{-1}\bX_{1}^\top\bX_{1}\Big) \\
& \geq \frac{||\theta_{1}-\theta_{2}||^{2}/\sigma^{2}}{\frac{1}{\lambda_{\min}(\bX_{1}^\top\bX_{1})}+\frac{1}{\lambda_{\min}(\bX_{2}^\top\bX_{2})}}
\end{split}
\end{equation*}
The first inequality uses the Rayleigh-Ritz theorem, and the second inequality uses the relation $\bY(\bX+\bY)^{-1}\bX=(\bX^{-1}+\bY^{-1})^{-1}$, where $\bX$ and $\bY$ are both invertible matrices. This relation can be derived by taking inverse on both sides of the equation $\bX^{-1}(\bX+\bY)\bY^{-1}=\bX^{-1}\bX\bY^{-1}+\bX^{-1}\bY\bY^{-1}=\bY^{-1}+\bX^{-1}$.

The results above show that given two datasets $\cH_{1}$ and $\cH_{2}$, the type-I error probability of the homogeneity test only depends on the selection of threshold $\upsilon$, while the type-II error probability also depends on the ground-truth parameters ($\theta_{1}$, $\theta_{2}$) and variance of noise $\sigma^{2}$. If either $\bX_{1}$ or $\bX_{2}$ is rank-deficient, the type-II error probability will be trivially upper bounded by $F(\upsilon;df,0)$, which means for a smaller upper bound of type-I error probability (i.e., $1-F(\upsilon;df,0)$), the upper bound of type-II error probability (i.e., $F(\upsilon;df,0)$) will be large. Intuitively, for a certain level of type-I error, to ensure a smaller type-II error probability in the worst case, we at least need both $\bX_{1}$ and $\bX_{2}$ to be rank-sufficient and the value of $\frac{||\theta_{1}-\theta_{2}||^{2}/\sigma^{2}}{\frac{1}{\lambda_{\min}(\bX_{1}^\top\bX_{1})}+\frac{1}{\lambda_{\min}(\bX_{2}^\top\bX_{2})}}$ to be large. Similar idea is also found in \cite{wu2018learning, gentile2014online, gentile2017context}, where they require the confidence bounds of the estimators (which correspond to the minimum eigenvalue of the correlation matrix in our case) to be small enough, w.r.t. $||\theta_{1}-\theta_{2}||^{2}$, to ensure their change detection or cluster identification is accurate. Here we unify the analysis of these two tasks with this homogeneity test.

These error probabilities are the key concerns in our problem: in change detection, they correspond to the early and late detection of change points \cite{wu2018learning}; and in cluster identification, they correspond to missing a user model in the neighborhood and placing a wrong user model in the neighborhood \cite{gentile2014online}. 
Given it is impossible to completely eliminate these two types of errors in a non-deterministic algorithm, the uniformly most powerful property of the test defined in Eq~\eqref{eq_chis_test} guarantees its sensitivity is optimal at any level of specificity.

\subsection{Algorithm}
In the environment specified in Section \ref{subsec:problemFormulation}, the user's reward mapping function is piecewise stationary (e.g., the line segments on each user's interaction trace in Figure \ref{fig:envAndAlgo}), which calls for the learner to actively detect changes and re-initialize the estimator to avoid distortion from outdated observations \cite{yu2009piecewise, cao2019nearly, besson2019generalized,wu2018learning}. A limitation of these methods is that they do not attempt to reuse outdated observations because they implicitly assume each stationary period has an unique parameter. Our setting relaxes this by allowing for existence of identical reward mappings across users and time (e.g., the orange line segments in Figure \ref{fig:envAndAlgo}), which urges the learner to take advantage of this situation by identifying and aggregating observations with the same parameter to obtain a more accurate reward estimation.

Since neither the change points nor the grouping structure is known, in order to reuse past observations while avoiding distortion, the learner needs to accurately detect change points, stores observations in the interval between two consecutive detection together, and then correctly identify intervals with the same parameter as the current one. In this paper, we propose to unify these two operations using the test in Section \ref{subsec:similarity_measure}, which leads to the algorithm Dynamic Clustering of Bandits, or \model{} in short. \model{} forms a two-level hierarchy as shown in Figure \ref{fig:envAndAlgo}: 
at the lower level, it stores observations in each interval and their sufficient statistics in a user model;
at the upper level, it detects change in user's reward function to decide when to create new user models and clusters individual user models for arm selection. Detailed steps of \model{} are explained in Algorithm \ref{algorithm}.

\begin{algorithm}[ht]
    \caption{Dynamic Clustering of Bandits (\model)}\label{algorithm}
  \begin{algorithmic}[1]
    \STATE \textbf{Input:} sliding window size $\tau$, $\delta, \delta_{e} \in (0,1)$, threshold for change detection and neighbor identification $\upsilon^{e}$ and $\upsilon^{c}$, and regularization parameter $\lambda$
    \STATE \textbf{Initialization:} for each user model $\bM_{i,0}, \forall i \in \cU$: $\bA_{i,0}=\textbf{0} \in \bR^{d \times d}$, $\bb_{i,0}=\textbf{0} \in \bR^{d}$, $\cH_{i,0}=\emptyset$, $\hat{e}_{i,0}=0$; the set of outdated user models $\bO_{0}=\emptyset$, and up-to-date user models $\bU_{0}=\left\{\bM_{i,0}\right\}_{i \in \cU}$
    \FOR{$t=1,2,...,T$}
        \STATE Observe user $i_{t} \in \cU$, and set of available arms $C_{t}=\{x_{t,1},..., x_{t,K}\}$
        \STATE Choose arm $\bx_{t} \in C_{t}$ by Eq \ref{eq:UCB}: $\argmax_{\bx \in C_{t}}{\bx^{\top}\hat{\theta}_{\hat{V}_{i_{t},t-1}}+CB_{\hat{V}_{i_{t},t-1}}(x)}$
        \STATE Observe reward $y_{t}$ from user $i_{t}$
        \STATE Compute $e_{i_{t},t}=\textbf{1}\left\{S(\cH_{i_{t},t-1},(\bx_{t}^{\top},y_{t}))>\upsilon^{e}\right\}$
        \STATE Update $\hat{e}_{i_{t},t}=\sum_{\tilde{t}_{i_{t}}(\tau) < j \leq t:i_{j}=i_{t}}e_{i_{t},j}$
        \IF {$\hat{e}_{i_{t},t}\leq 1-F(\upsilon^{e};1,0) + \sqrt{\frac{\log{1/\delta_{e}}}{2\tau}}$}
            \IF{$e_{i_{t},t}=0$}
                \STATE $\bM_{i_{t},t}$: $\cH_{i_{t},t}=\cH_{i_{t},t-1}\cup (\bx_{t},y_{t})$, $\bA_{i_{t},t}=\bA_{i_{t},t-1}+\bx_{t}\bx_{t}^{\top}$, $\bb_{i_{t},t}=\bb_{i_{t},t-1}+\bx_{t}y_{t}$
            \ENDIF
        \ELSE
            \STATE $\bO_{t}=\bO_{t-1} \cup \bM_{i_{t},t-1}$, $\hat{e}_{i_{t},t}=0$
            \STATE Replace $\bM_{i_{t},t-1}$ with $\bM_{i_{t},t}=\left(A_{i_{t},t}=\textbf{0}, b_{i_{t},t}=\textbf{0}, \cH_{i,t}=\emptyset\right)$ in $\bU_{t}$
        \ENDIF
        \STATE Update user $i_{t}$'s neighborhood:  $\hat{V}_{i_{t},t}=\left\{\bM \in \bU_{t}\cup\bO_{t}:S(\cH_{i_{t},t},\cH) \leq \upsilon^{c}\right\}$
    \ENDFOR
  \end{algorithmic}
\end{algorithm}

The lower level of \model{} manages observations associated with each user $i \in \cU$ in user models, denoted by $\bM_{i,t}$. Each user model $\bM_{i,t}=(\bA_{i,t}, \bb_{i,t}, \cH_{i,t})$ stores:
\begin{enumerate}
    \item $\cH_{i,t}$: a set of observations associated with user $i$ since the initialization of $\bM_{i,t}$ up to time $t$, where each element is a context vector and reward pair $(\bx_{k}, y_{k})$.
    \item Sufficient statistics: $\bA_{i,t}=\sum_{(\bx_{k},\cdot) \in \cH_{i,t}}\bx_{k}\bx_{k}^{\top}$ and $\bb_{i,t}=\sum_{(\bx_{k},y_{k}) \in \cH_{i,t}}\bx_{k}y_{k}$.
\end{enumerate}

Every time \model{} detects change in a user's reward mapping function, a new user model is created to replace the previous one (line 15 in Algorithm \ref{algorithm}). We refer to the replaced user models as outdated models and the others up-to-date ones.
We denote the set of all outdated user models at time $t$ as $\bO_{t}$ and the up-to-date ones as $\bU_{t}$. In Figure \ref{fig:envAndAlgo}, the row of circles next to $\bM_{1,t-1}$ represents all the user models for user $1$, red ones denote outdated models and the blue one denotes up-to-date model. 

The upper level of \model{} is responsible for managing the user models via change detection and model clustering. It replaces outdated models in each user and aggregates models across users and time for arm selection.


\noindent\textbf{$\bullet$ Change detection.} 
A one-sample homogeneity test is used to construct a test variable $e_{i_{t},t}=\textbf{1}\left\{s(\cH_{i_{t},t-1},\left\{(\bx_{t},y_{t})\right\})>\upsilon^{e}\right\}$ to measure whether the user model $\bM_{i_{t},t-1}$ is `admissible' to the new observation $(\bx_{t},y_{t})$. $\upsilon^{e}$ is a chosen threshold for change detection.
To make more reliable change detection, we use the empirical mean of $e_{i_{t},t}$ in a sliding window of size $\min(|\cH_{i_{t},t-1}|,\tau)$ as the test statistic, denoted as $\hat{e}_{i_{t},t}=\frac{1}{\min(|\cH_{i_{t},t-1}|,\tau)}\sum_{k}e_{i_{t},k}$.
Lemma \ref{lem:earlyDetection} specifies the upper bound of early detection probability using $\hat{e}_{i,t}$, which is used for selecting threshold for it.
\begin{lemma}\label{lem:earlyDetection}
From Lemma \ref{lem:type1Similarity}, type-1 error probability $P(e_{i,t}=1)\leq 1-F(\upsilon^{e};1,0)$, and thus $\mathbb{E}[e_{i,t}]\leq 1-F(\upsilon^{e};1,0)$.  Applying Hoeffding inequality gives,
\begin{equation*}
    P\Big(\hat{e}_{i,t}>1-F(\upsilon^{e};1,0) + \sqrt{\frac{\log{1/\delta_{e}}}{2\tau}}\Big) \leq \delta_{e}
\end{equation*}
\end{lemma}
At any time step $t$, \model{} only updates $\bM_{i_{t},t-1}$ when $e_{i_{t},t}=0$ (line 10-12 in Algorithm \ref{algorithm}). This guarantees that if the underlying reward distribution has changed, with a high probability we have $e_{i_{t},t}=1$, and thus the user model $\bM_{i_{t},t-1}$ will not be updated. 
This prevents any distortion in $\cH_{i_{t},t}$ by observations from different reward distributions.

When $\hat{e}_{i_{t},t}$ exceeds the threshold specified by Lemma \ref{lem:earlyDetection}, 
\model{} will inform the lower level to move $\bM_{i_{t},t-1}$ to the outdated model set $\bO_{t}=\bO_{t-1}\cup\left\{\bM_{i_{t},t-1}\right\}$; and then create a new model $\bM_{i_{t},t}=\left(A_{i_{t},t}=\textbf{0}, b_{i_{t},t}=\textbf{0}, \cH_{i,t}=\emptyset\right)$ for user $i_{t}$ as shown in line 13-16 in Algorithm \ref{algorithm}.


\noindent\textbf{$\bullet$ Clustering of user models.} 
In this step, \model{} finds the set of ``neighborhood'' user models $\hat{V}_{i_{t},t}$ of current user model $\bM_{i_{t}},t$, where $\hat{V}_{i_{t},t-1}=\left\{\bM=(\bA,\bb,\cH) \in \bU_{t}\cup\bO_{t}:s(\cH_{i_{t},t},\cH) \leq \upsilon^{c}\right\}$. Basically, \model{} executes homogeneity test between $\bM_{i_{t},t}$ and all other user models $\bM \in \bU_{t}\cup\bO_{t}$ (both outdated and up-to-date) with a given threshold $\upsilon^{c}$ (line 17 in Algorithm \ref{algorithm}). Lemma \ref{lem:type1Similarity} and \ref{lem:type2Similarity} again specify error probabilities of each decision.


When selecting an arm for user $i_{t}$ at time $t$, \model{} aggregates the sufficient statistics of user models in neighborhood $\hat{V}_{i_{t},t-1}$. Then it adopts the popular UCB strategy \cite{auer2002using,li2010contextual} to balance exploitation and exploration. Specifically, \model{} selects arm $\bx_{t}$ that maximizes the UCB score computed by aggregated sufficient statistics as follows (line 5 in Algorithm \ref{algorithm}),
\begin{equation}\label{eq:UCB}
    \bx_{t}=\argmax_{\bx \in C_{t}}{\bx^{\top}\hat{\theta}_{\hat{V}_{i_{t},t-1}}+CB_{\hat{V}_{i_{t},t-1}}(\bx)}
\end{equation}
In Eq \eqref{eq:UCB}, $\hat{\theta}_{\hat{V}_{i_{t},t-1}}=\bA_{\hat{V}_{i_{t},t-1}}^{-1}\bb_{\hat{V}_{i_{t},t-1}}$ is the ridge regression estimator using aggregated statistics $\bA_{\hat{V}_{i_{t},t-1}}=\lambda \bI_{d}+\sum_{(\bA_{j},\bb_{j},\cH_{j}) \in \hat{V}_{i_{t},t-1}}\bA_{j}$ and $\bb_{\hat{V}_{i_{t},t-1}}=\sum_{(\bA_{j},\bb_{j},\cH_{j}) \in \hat{V}_{i_{t},t-1}}\bb_{j}$; the confidence bound of reward estimation for arm $\bx$ is $CB_{\hat{V}_{i_{t},t-1}}(\bx)=\alpha_{\hat{V}_{i_{t},t-1}}\sqrt{\bx^{\top} \bA_{\hat{V}_{i_{t},t-1}}^{-1}\bx}$, where $\alpha_{\hat{V}_{i_{t},t-1}}=\sigma\sqrt{d\log{(1+\frac{\sum_{(A_{j},b_{j},\cH_{j}) \in \hat{V}_{i_{t},t-1}}|\cH_{j}|}{d\lambda})} + 2\log{\frac{1}{\delta}}}+\sqrt{\lambda}$.

\subsection{Regret analysis}\label{subsec:regretAnalysis}
Denote $R_{T}=\sum_{t=1}^{T} \theta_{i_{t}}^{\top} \bx_{t}^{*} - \theta_{i_{t}}^{\top} \bx_{t}$ as the accumulative regret, where $\bx_{t}^{*}=\argmax_{x_{t,j} \in C_{t}}\theta_{i_{t}}^{\top} \bx_{t,j}$ is the optimal arm at time $t$. Our regret analysis relies on the high probability results in \cite{abbasi2011improved} and decomposition of "good" and "bad" events according to change detection and clustering results. Specifically, the "good" event corresponds to the case where the aggregated statistics in Eq \ref{eq:UCB} contains and only contains all the past observations associated with the ground truth unique bandit parameter at current time step. The "bad" event is simply the complement of "good" event that occurs as a result of errors in change detection and clustering, and we can analyze their occurrences based on our results in Lemma \ref{lem:type1Similarity} and Lemma \ref{lem:type2Similarity}. 
Following this idea,  the upper regret bound of \model{} given in Theorem \ref{thm:regretBound} can be derived. The full proof, along with ancillary results, are provided in the appendix. 

\begin{theorem}\label{thm:regretBound}
Under Assumptions \ref{assump1}, \ref{assump2} and \ref{assump3}, the  regret of \model{} is upper bounded by:

\begin{equation*}
 R_{T} = O\Big(\sigma d\sqrt{T\log^{2}{T}}\big(\sum_{k=1}^{m}\sqrt{p_{k}}\big)+\sum_{i \in \cU} \Gamma_{i}(T) \cdot C \Big)
\end{equation*}

where $C=\frac{1}{1-\delta^{e}}+\frac{\sigma^{2}}{\gamma^{2}\lambda'^{2}}\log{\frac{d}{\delta^{'}}}$, 
with a probability at least $(1-\delta)(1-\frac{\delta_{e}}{1-\delta_{e}})(1-\delta^{'})$.
\end{theorem}


This upper regret bound depends both on the intrinsic clustering and non-stationary property of user set $\cU$. Since the proposed \model{} algorithm addresses a bandit problem that generalizes environment settings in several existing studies, here we discuss how \model{} compares with these algorithms when reduced to their corresponding environment settings.


\textbf{Case 1}: Setting $m=1$, $n=1$ and $\Gamma_{1}(T)=1$ reduces the problem to the basic linear bandit setting, because the environment consists of only one user with a stationary reward distribution for the entire time of interaction. With only one user who has a stationary reward distribution, we have $\sum_{k=1}^{1}\sqrt{p_{k}}=1$ where $p_{k}$ is frequency of occurrences of $\phi_{k}$ in $T$ as defined in Section \ref{subsec:problemFormulation}. In addition, since there is only one stationary period, the added regret caused by late detection does not exist; and the added regret due to the failure in clustering can be bounded by a constant as later shown in Lemma \ref{lem:thirdTerm}, which only depends on environment variables. The upper regret bound of \model{} then becomes $O\big(\sigma d\sqrt{T\log^{2}{T}}\big)$, which achieves the same order of regret as that in LinUCB \cite{abbasi2011improved}. 

\textbf{Case 2}: Setting $\Gamma_{i}(T)=1, \forall i \in \cU$ reduces the problem to the online clustering of bandit setting \cite{gentile2014online}, because all users in the environment have a stationary reward distribution of their own. Similar to Case 1, the added regret caused by late detection becomes zero and the added regret due to the failure in clustering is bounded by a constant, which leads to the upper regret bound of $O\big(\sigma d\sqrt{T\log^{2}{T}}(\sum_{k=1}^{m}\sqrt{p_{k}}\big)$. \model{} achieves the same order of regret as that in CLUB \cite{gentile2014online}. 

\textbf{Case 3}: Setting $n=1$ reduces the problem to a piecewise stationary bandit setting, because the environment consists of only one user with piecewise stationary reward distributions. For the convenience of comparison, we can rewrite the upper regret bound of \model{} in the form of $O\big(\sum_{k \in [m]}R_{Lin}(|N^{\phi}_{k}(T)|)+\Gamma_{1}(T)\big)$, where $R_{Lin}(t)=O\Big(d\sqrt{t\log^{2}{t}}\Big)$  \cite{abbasi2011improved} and $N^{\phi}_{k}(T)=\left\{1\leq t^{'} \leq T: \theta_{i_{t^{'}},t^{'}}=\phi_{k}\right\}$ is the set of time steps up to time $T$ when the user being served has the bandit parameter equal to $\phi_{k}$. Detailed derivation of this is given in Section \ref{subsec:proofEq3}. Note that the upper regret bound of dLinUCB \cite{wu2018learning} for this setting is $O\big(\Gamma_{1}(T)R_{Lin}(S_{max})+\Gamma_{1}(T)\big)$, where $S_{max}$ denotes the maximum length of stationary periods. The regret of \model{} depends on the number of unique bandit parameters in the environment, instead of the number of stationary periods as in dLinUCB, because \model{} can reuse observations from previous stationary periods. This suggests \model{} has a lower regret bound if different stationary periods share the same unique bandit parameters; for example, in situations where a future reward mapping function switches back to a previous one.

\section{Experiments}
We investigate the empirical performance of the proposed algorithm by comparing with a list of state-of-the-art baselines for both non-stationary bandits and clustered bandits on synthetic and real-world recommendation datasets.

\subsection{Experiment setup and baselines}
\noindent\textbf{$\bullet$ Synthetic dataset.} We create a set of unique bandit parameters $\left\{\phi_{k}\right\}_{k=1}^{m}$ and arm pool $\left\{\bx_{j}\right\}_{j=1}^{K}$ ($K=1000$), where $\phi_{k}$ and $\bx_{j}$ are first sampled from $N(\mathbf{0}_d,\bI_{d})$ with $d=25$ and then normalized so that $\forall k,j, \lVert \phi_{k} \rVert = 1 \text{ and } \lVert \bx_{j} \rVert = 1$. When sampling $\left\{\phi_{k}\right\}_{k=1}^{m}$, the separation margin $\gamma$ is set to 0.9 and enforced via rejection sampling. $n$ users are simulated. In each user, we sample a series of time intervals from $(S_{min},S_{max})$ uniformly; and for each time interval, we sample a unique parameter from $\left\{\phi_{k}\right\}_{k=1}^{m}$ as the ground-truth bandit parameter for this period. This creates asynchronous changes and clustering structure in users' reward functions. The users are served in a round-robin fashion. At time step $t=1, 2, \dots, T$, a subset of arms are randomly chosen and disclosed to the learner. Reward of the selected arm is generated by the linear function governed by the corresponding bandit parameter and context vector, with additional Gaussian noise sampled from $N(0,\sigma^{2})$.

\noindent\textbf{$\bullet$ LastFM dataset.} The LastFM dataset is extracted from the music streaming service Last.fm \cite{cesa2013gang}, which contains 1892 users and 17632 items (artists). `Listened artists' of each user are treated as positive feedback. We followed \cite{wu2018learning} to preprocess the dataset and  simulate a clustered non-stationary environment by creating 20 `hybrid users'. 
We first discard users with less than 800 observations and then use PCA to reduce the dimension of TF-IDF feature vector to $d=25$.
We create hybrid users by sampling three real users uniformly and then concatenating their associated data points together. Hence, data points of the same real user would appear in different hybrid users, which is analogous to stationary periods that share the same unique bandit parameters across different users and time.

\noindent\textbf{$\bullet$ Baselines.}
We compare \model{} with a set of state-of-the-art bandit algorithms: linear bandit LinUCB \cite{abbasi2011improved}, non-stationary bandit dLinUCB \cite{wu2018learning} and adTS \cite{slivkins2008adapting}, as well as online clustering bandit CLUB \cite{gentile2014online}. In simulation based experiments, we also include oracle-LinUCB for comparison, which runs an instance of LinUCB for each unique bandit parameter. Comparing with it helps us understand the added regret from errors in change detection and clustering. 

\subsection{Experiment results}

\begin{figure*}[ht]
\centerline{\includegraphics[width=\linewidth]{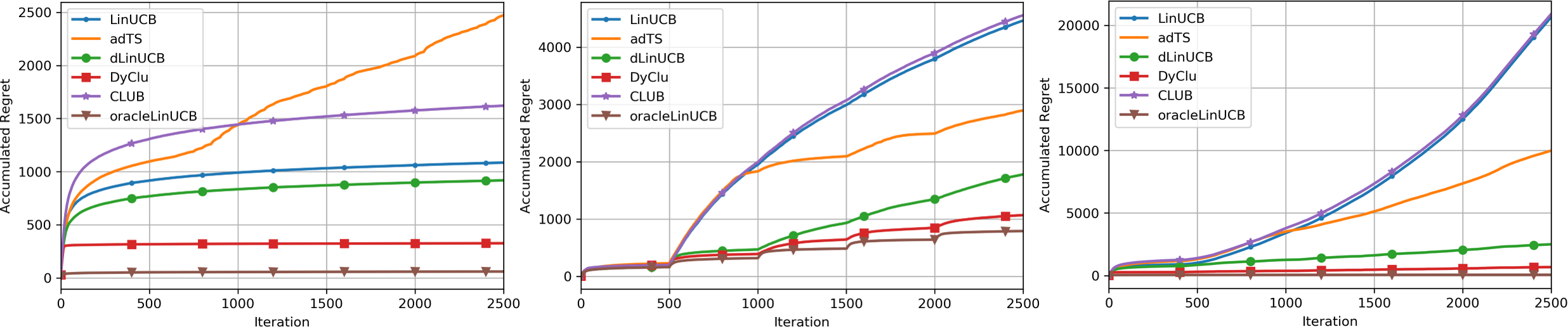}}
\caption{Accumulated regret on synthetic datasets with three different environment settings. Environment 1: $n=100$ users share a global set of $m=5$ unique bandit parameters, and each user remains stationary all the time. Environment 2: $n=20$ user with fixed stationary period length $500$; each period sample a unique bandit parameter. Environment 3: $n=100$ users share a global set of $m=5$ unique bandit parameters, and each user changes in a asynchronous manner.}
\label{fig:simulationEnvs}
\end{figure*}

\noindent\textbf{$\bullet$ Empirical comparisons on synthetic dataset.} We compare accumulated regret of all bandit algorithms under three environment settings, and the results are reported in Figure \ref{fig:simulationEnvs}. Environment 1 simulates the online clustering setting in \cite{gentile2014online}, where \emph{no} change in the reward function is introduced. \model{} outperformed other baselines, including CLUB, demonstrating the quality of its identified clustering structure. Specifically, compared with adTS that incurs high regret as a result of too many false detections, the change detection in \model{} and dLinUCB has much less false positives, as there is no change in each user's reward distribution. 
Environment 2 simulates the piecewise stationary setting in \cite{wu2018learning}. Algorithms designed for stationary environment, e.g., CLUB and LinUCB suffer from a linear regret after the first change point. \model{} achieved the best performance, with a wide margin with second best, dLinUCB, which is designed for this environment. 
Environment 3 combines previous two settings with both non-stationarity and clustering structure. \model{} again outperformed others. It is worth noting that regret of all algorithms increased compared with Environment 1 due to the nonstationarity, but the increase in \model{} is the smallest. And in all settings, \model{}'s performance is closest to the oracle LinUCB's, which shows that \model{} can correctly cluster and aggregate observations from the dynamically changing users.

\noindent\textbf{$\bullet$ Sensitivity to environment settings.}
According to our regret analysis, the performance of \model{} depends on environment parameters like the number of unique bandit parameters $m$, the number of stationary periods $\Gamma_{i}(T)$ for $i\in \cU$, and variance of Gaussian noise $\sigma^{2}$. We investigate their influence on \model{} and baselines, by varying these parameters while keeping the others fixed. The accumulated regret under different settings are reported in Table \ref{tb:simulationResults}. \model{} outperformed other baselines in all 9 different settings, and the changes of its regret align with our theoretical analysis. A larger number of unique parameters $m$ leads to higher regret of \model{} as shown in setting 1, 2 and 3, since observations are split into more clusters with smaller size each. In addition, larger number of stationary periods incurs more errors in change detection, leading to an increased regret. This is confirmed by results in setting 4, 5 and 6. Lastly, as shown in setting 7, 8 and 9, larger Gaussian noise leads to a higher regret, as it slows down convergence of reward estimation and change detection.

\begin{table*}[ht]
\centering
\caption{Comparison of accumulated regret under different environment settings.}
\begin{tabular}{p{0.2cm} p{0.3cm} p{0.3cm} p{0.45cm} p{0.45cm} p{0.45cm} p{0.45cm} p{0.95cm} p{0.95cm} p{0.95cm} p{0.95cm} p{0.95cm} p{0.95cm}}
\toprule
& n & m & $S_{min}$ & $S_{max}$ & T & $\sigma$ & oracle. & LinUCB & adTS & dLin. & CLUB & DyClu \\
\cmidrule(r{4pt}){2-7} \cmidrule(l){8-13}
1 & 100 & 10 & 400 & 2500 & 2500 & 0.09 & 115 & 19954 & 9872 & 2432 & 20274 & 853\\
2 & 100 & 50 & 400 & 2500 & 2500 & 0.09 & 489 & 20952 & 9563 & 2420 & 21205 & 1363 \\
3 & 100 & 100 & 400 & 2500 & 2500 & 0.09 & 873 & 21950 & 10961 & 2549 & 22280 & 1958\\
4 & 100 & 10 & 200 & 400 & 2500 & 0.09 & 112 & 39249 & 36301 & 10831 & 39436 & 3025 \\
5 & 100 & 10 & 800 & 1000 & 2500 & 0.09 & 113 & 34385 & 13788 & 3265 & 34441 & 1139\\
6 & 100 & 10 & 1200 & 1400 & 2500 & 0.09 & 112 & 24769 & 8124 & 2144 & 24980 & 778 \\
7 & 100 & 10 & 400 & 2500 & 2500 & 0.12 & 166 & 22453 & 10567 & 3301 & 22756 & 1140\\
8 & 100 & 10 & 400 & 2500 & 2500 & 0.15 & 232 & 19082 & 10000 & 5872 & 19427 & 1487\\
9 & 100 & 10 & 400 & 2500 & 2500 & 0.18 & 307 & 23918 & 11255 & 9848 & 24050 & 1956\\
\bottomrule
\end{tabular}
\label{tb:simulationResults}
\end{table*}

\begin{figure*}[ht]
    \begin{center}
    \centerline{\includegraphics[width=0.5\linewidth]{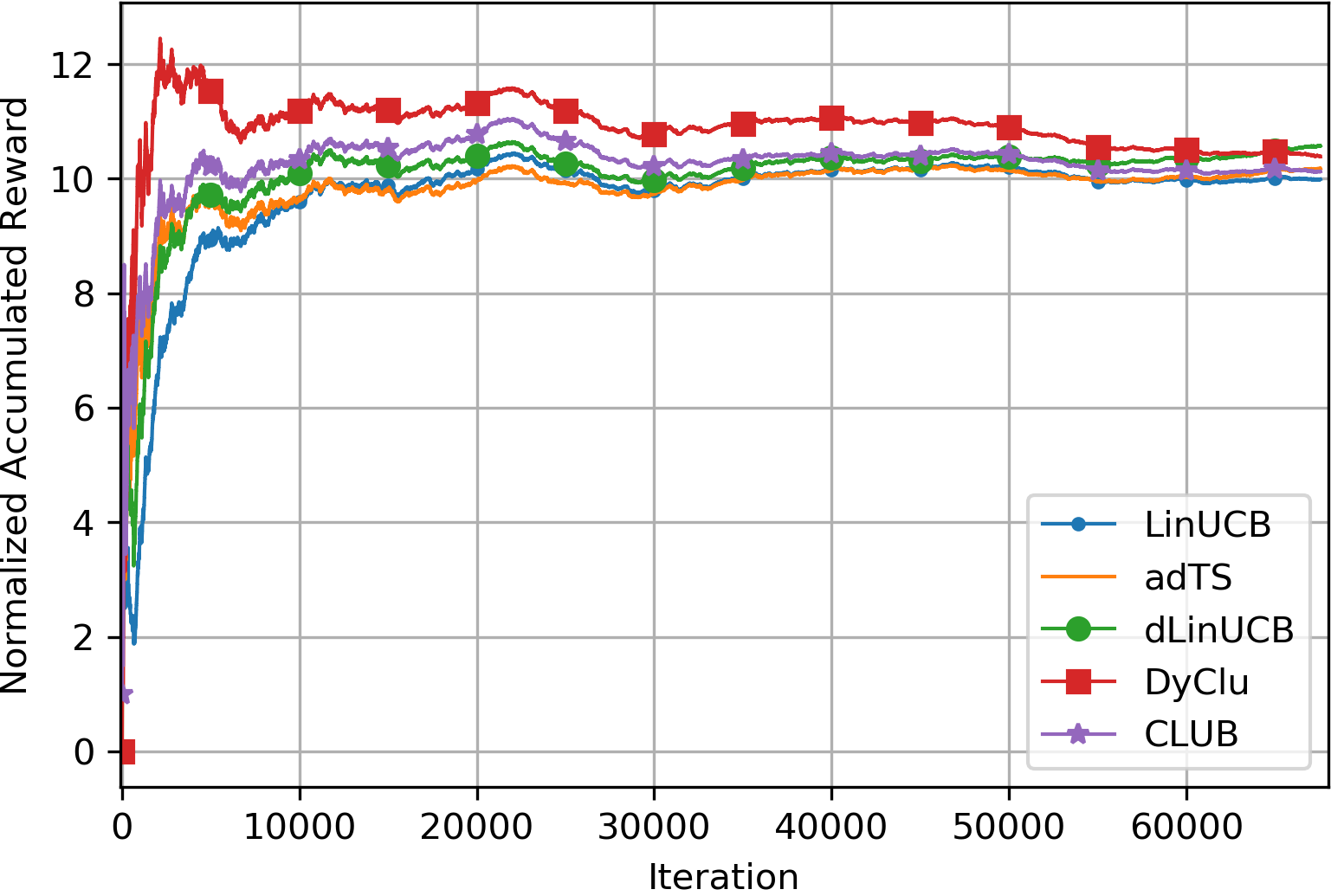}}
    \caption{Comparison of accumulated reward normalized by a random policy on LastFM dataset.}
    \label{fig:LastFMexp}
    \end{center}
    \vspace{-4mm}
\end{figure*}

\noindent\textbf{$\bullet$ Empirical comparisons on LastFM.}
We report normalized accumulative reward (ratio between baselines and uniformly random arm selection strategy \cite{wu2019dynamic}) on LastFM in Figure \ref{fig:LastFMexp}. In this environment, realizing both non-stationarity and clustering structure is important for an online learning algorithm to perform. \model{}'s improvement over other baselines confirms its quality in partitioning and aggregating relevant data points across users. 
The advantage of \model{} is more apparent at the early stage of learning, where each local user model has not collected sufficient amount of observations for individualized reward estimation; and thus change detection and clustering are more difficult there.


\section{Conclusion}
In this work, we unify the efforts in non-stationary bandits and online clustering of bandits via homogeneity test. The proposed solution adaptively detects changes in the underlying reward distribution and clusters bandit models for aggregated arm selection. The resulting upper regret bound matches with the ideal algorithm's only up to a constant; and extensive empirical evaluations validate its effectiveness in learning in a non-stationary and clustered environment. 

There are still several directions left open in this research. Currently, both change detection and cluster identification are performed at a user level; it can be further contextualized at the arm level \cite{wu2019dynamic,li2016collaborative}. Despite the existence of multiple users, all computations are done in a centralized manner; to make the solution more practical, asynchronized and distributed model update is more desired. 

\newpage
\bibliography{bibfile}

\begin{thebibliography}{29}
\providecommand{\natexlab}[1]{#1}
\providecommand{\url}[1]{\texttt{#1}}
\expandafter\ifx\csname urlstyle\endcsname\relax
  \providecommand{\doi}[1]{doi: #1}\else
  \providecommand{\doi}{doi: \begingroup \urlstyle{rm}\Url}\fi

\bibitem[Abbasi-Yadkori et~al.(2011)Abbasi-Yadkori, P{\'a}l, and
  Szepesv{\'a}ri]{abbasi2011improved}
Yasin Abbasi-Yadkori, D{\'a}vid P{\'a}l, and Csaba Szepesv{\'a}ri.
\newblock Improved algorithms for linear stochastic bandits.
\newblock In \emph{Advances in Neural Information Processing Systems}, pages
  2312--2320, 2011.

\bibitem[Auer(2002)]{auer2002using}
Peter Auer.
\newblock Using confidence bounds for exploitation-exploration trade-offs.
\newblock \emph{Journal of Machine Learning Research}, 3\penalty0
  (Nov):\penalty0 397--422, 2002.

\bibitem[Besson and Kaufmann(2019)]{besson2019generalized}
Lilian Besson and Emilie Kaufmann.
\newblock The generalized likelihood ratio test meets klucb: an improved
  algorithm for piece-wise non-stationary bandits.
\newblock \emph{arXiv preprint arXiv:1902.01575}, 2019.

\bibitem[Buccapatnam et~al.(2013)Buccapatnam, Eryilmaz, and
  Shroff]{buccapatnam2013multi}
Swapna Buccapatnam, Atilla Eryilmaz, and Ness~B Shroff.
\newblock Multi-armed bandits in the presence of side observations in social
  networks.
\newblock In \emph{52nd IEEE Conference on Decision and Control}, pages
  7309--7314. IEEE, 2013.

\bibitem[Cantrell et~al.(1991)Cantrell, Burrows, and
  Vuong]{cantrell1991interpretation}
R~Stephen Cantrell, Peter~M Burrows, and Quang~H Vuong.
\newblock Interpretation and use of generalized chow tests.
\newblock \emph{International Economic Review}, pages 725--741, 1991.

\bibitem[Cao et~al.(2019)Cao, Zheng, Kveton, and Xie]{cao2019nearly}
Yang Cao, Wen Zheng, Branislav Kveton, and Yao Xie.
\newblock Nearly optimal adaptive procedure for piecewise-stationary bandit: a
  change-point detection approach.
\newblock \emph{AISTATS,(Okinawa, Japan)}, 2019.

\bibitem[Cesa-Bianchi et~al.(2013)Cesa-Bianchi, Gentile, and
  Zappella]{cesa2013gang}
Nicolo Cesa-Bianchi, Claudio Gentile, and Giovanni Zappella.
\newblock A gang of bandits.
\newblock In \emph{Advances in Neural Information Processing Systems}, pages
  737--745, 2013.

\bibitem[Chen et~al.(2019)Chen, Lee, Luo, and Wei]{chen2019new}
Yifang Chen, Chung-Wei Lee, Haipeng Luo, and Chen-Yu Wei.
\newblock A new algorithm for non-stationary contextual bandits: Efficient,
  optimal, and parameter-free.
\newblock \emph{arXiv preprint arXiv:1902.00980}, 2019.

\bibitem[Chow(1960)]{chow1960tests}
Gregory~C Chow.
\newblock Tests of equality between sets of coefficients in two linear
  regressions.
\newblock \emph{Econometrica: Journal of the Econometric Society}, pages
  591--605, 1960.

\bibitem[Chu et~al.(2011)Chu, Li, Reyzin, and Schapire]{chu2011contextual}
Wei Chu, Lihong Li, Lev Reyzin, and Robert Schapire.
\newblock Contextual bandits with linear payoff functions.
\newblock In \emph{Proceedings of the Fourteenth International Conference on
  Artificial Intelligence and Statistics}, pages 208--214, 2011.

\bibitem[Filippi et~al.(2010)Filippi, Cappe, Garivier, and
  Szepesv{\'a}ri]{filippi2010parametric}
Sarah Filippi, Olivier Cappe, Aur{\'e}lien Garivier, and Csaba Szepesv{\'a}ri.
\newblock Parametric bandits: The generalized linear case.
\newblock In \emph{Advances in Neural Information Processing Systems}, pages
  586--594, 2010.

\bibitem[Garivier and Moulines(2011)]{garivier2011on}
Aur\'{e}lien Garivier and Eric Moulines.
\newblock On upper-confidence bound policies for switching bandit problems.
\newblock In \emph{Proceedings of the 22nd International Conference on
  Algorithmic Learning Theory}, ALT'11, pages 174--188, Berlin, Heidelberg,
  2011. Springer-Verlag.
\newblock ISBN 9783642244117.

\bibitem[Gentile et~al.(2014)Gentile, Li, and Zappella]{gentile2014online}
Claudio Gentile, Shuai Li, and Giovanni Zappella.
\newblock Online clustering of bandits.
\newblock In \emph{International Conference on Machine Learning}, pages
  757--765, 2014.

\bibitem[Gentile et~al.(2017)Gentile, Li, Kar, Karatzoglou, Zappella, and
  Etrue]{gentile2017context}
Claudio Gentile, Shuai Li, Purushottam Kar, Alexandros Karatzoglou, Giovanni
  Zappella, and Evans Etrue.
\newblock On context-dependent clustering of bandits.
\newblock In \emph{Proceedings of the 34th International Conference on Machine
  Learning-Volume 70}, pages 1253--1262. JMLR. org, 2017.

\bibitem[Hariri et~al.(2015)Hariri, Mobasher, and Burke]{hariri2015adapting}
Negar Hariri, Bamshad Mobasher, and Robin Burke.
\newblock Adapting to user preference changes in interactive recommendation.
\newblock In \emph{Twenty-Fourth International Joint Conference on Artificial
  Intelligence}, 2015.

\bibitem[Hartland et~al.(2006)Hartland, Gelly, Baskiotis, Teytaud, and
  Sebag]{hartland2006multi}
C{\'e}dric Hartland, Sylvain Gelly, Nicolas Baskiotis, Olivier Teytaud, and
  Michele Sebag.
\newblock Multi-armed bandit, dynamic environments and meta-bandits, 2006.
\newblock In \emph{NIPS-2006 workshop, Online trading between exploration and
  exploitation, Whistler, Canada}, 2006.

\bibitem[Kleinberg et~al.(2008)Kleinberg, Slivkins, and
  Upfal]{kleinberg2008multi}
Robert Kleinberg, Aleksandrs Slivkins, and Eli Upfal.
\newblock Multi-armed bandits in metric spaces.
\newblock In \emph{Proceedings of the fortieth annual ACM symposium on Theory
  of computing}, pages 681--690, 2008.

\bibitem[Li et~al.(2010)Li, Chu, Langford, and Schapire]{li2010contextual}
Lihong Li, Wei Chu, John Langford, and Robert~E Schapire.
\newblock A contextual-bandit approach to personalized news article
  recommendation.
\newblock In \emph{Proceedings of the 19th international conference on World
  wide web}, pages 661--670. ACM, 2010.

\bibitem[Li et~al.(2016)Li, Karatzoglou, and Gentile]{li2016collaborative}
Shuai Li, Alexandros Karatzoglou, and Claudio Gentile.
\newblock Collaborative filtering bandits.
\newblock In \emph{Proceedings of the 39th International ACM SIGIR conference
  on Research and Development in Information Retrieval}, pages 539--548. ACM,
  2016.

\bibitem[Li et~al.(2019)Li, Chen, and Leung]{li2019improved}
Shuai Li, Wei Chen, and Kwong-Sak Leung.
\newblock Improved algorithm on online clustering of bandits.
\newblock \emph{arXiv preprint arXiv:1902.09162}, 2019.

\bibitem[Maillard and Mannor(2014)]{maillard2014latent}
Odalric-Ambrym Maillard and Shie Mannor.
\newblock Latent bandits.
\newblock In \emph{International Conference on Machine Learning}, pages
  136--144, 2014.

\bibitem[Russac et~al.(2019)Russac, Vernade, and Capp{\'e}]{russac2019weighted}
Yoan Russac, Claire Vernade, and Olivier Capp{\'e}.
\newblock Weighted linear bandits for non-stationary environments.
\newblock In \emph{Advances in Neural Information Processing Systems}, pages
  12017--12026, 2019.

\bibitem[Siegmund(2013)]{siegmund2013sequential}
David Siegmund.
\newblock \emph{Sequential analysis: tests and confidence intervals}.
\newblock Springer Science \& Business Media, 2013.

\bibitem[Slivkins and Upfal(2008)]{slivkins2008adapting}
Aleksandrs Slivkins and Eli Upfal.
\newblock Adapting to a changing environment: the brownian restless bandits.
\newblock In \emph{COLT}, pages 343--354, 2008.

\bibitem[Wilson(1978)]{wilson1978chow}
AL~Wilson.
\newblock When is the chow test ump?
\newblock \emph{The American Statistician}, 32\penalty0 (2):\penalty0 66--68,
  1978.

\bibitem[Wu et~al.(2016)Wu, Wang, Gu, and Wang]{wu2016contextual}
Qingyun Wu, Huazheng Wang, Quanquan Gu, and Hongning Wang.
\newblock Contextual bandits in a collaborative environment.
\newblock In \emph{Proceedings of the 39th International ACM SIGIR conference
  on Research and Development in Information Retrieval}, pages 529--538. ACM,
  2016.

\bibitem[Wu et~al.(2018)Wu, Iyer, and Wang]{wu2018learning}
Qingyun Wu, Naveen Iyer, and Hongning Wang.
\newblock Learning contextual bandits in a non-stationary environment.
\newblock In \emph{The 41st International ACM SIGIR Conference on Research \&
  Development in Information Retrieval}, pages 495--504. ACM, 2018.

\bibitem[Wu et~al.(2019)Wu, Wang, Li, and Wang]{wu2019dynamic}
Qingyun Wu, Huazheng Wang, Yanen Li, and Hongning Wang.
\newblock Dynamic ensemble of contextual bandits to satisfy users' changing
  interests.
\newblock In \emph{The World Wide Web Conference}, pages 2080--2090, 2019.

\bibitem[Yu and Mannor(2009)]{yu2009piecewise}
Jia~Yuan Yu and Shie Mannor.
\newblock Piecewise-stationary bandit problems with side observations.
\newblock In \emph{Proceedings of the 26th Annual International Conference on
  Machine Learning}, pages 1177--1184. ACM, 2009.

\end{thebibliography}

\newpage
\appendix

\section{Notations used in this paper}
Here we list the notations used in this paper and their descriptions in Table \ref{tb:notations}.

\begin{table*}[h]
\centering
\caption{Notations used in this paper.}
\begin{tabular}{p{0.2cm} p{1.5cm} p{12cm}}
\toprule
& Notation & Description \\
\cmidrule(r{4pt}){1-3} 
1 & $\cU$ & the set of all users, with its cardinality denoted by $n$ \\
2 & $C_{t}$ & the set of available arms at time $t$, with its cardinality denoted by $K$ \\
3 & $(x_{t},y_{t})$ & context vector of the selected arm and its observed reward at time $t$\\
4 & $\theta_{i,t}$ & bandit parameter of user $i$ at time $t$  \\
5 & $d$ & dimension of context vector and bandit parameter \\
6 & $\left\{\phi_{k}\right\}_{k=1}^{m}$ & a set of $m$ unique bandit parameters \\
7 & $k_{i,t}$ & index of unique bandit parameter associated with user $i$ at time $t$\\
8 & $\eta_{t}$ & Gaussian noise in the reward observed at time $t$ \\
9 & $\sigma^{2}$ & variance of the Gaussian noise in reward \\
10 & $\cN_{i}(T)$ & the set of time steps user $i \in \cU$ is served up to time $T$ \\
11 & $\Gamma_{i}(T)$ & the total number of stationary periods in $\cN_{i}(T)$ \\
12 & $c_{i,j}$ & the time step when the $j$'th change point of user $i$ occurs\\
13 & $N^{\phi}_{k}(t)$ & the set of time steps up to time $t$ that $\phi_{k}$ occurs \\
14 & $p_{k}$ & frequency of total occurrences of $\phi_{k}$\\
15 & $\cH$ & a set of observations \\
16 & $\bX,\by$ & design matrix and feedback vector of $\cH$\\
17 & ${\vartheta}$ & MLE on dataset $\cH$ \\
18 & $s(\cH_{1},\cH_{2})$ & homogeneity test statistic between $\cH_{1}$ and $\cH_{2}$ \\
19 & $\chi^{2}(df, \psi)$ & noncentral $\chi^{2}$ distribution with degree-of-freedom $df$ and noncentrality parameter $\psi$\\
20 & $F(\upsilon;df,\psi)$ & the cumulative density function of $\chi^{2}(df,\psi)$ evaluated at $\upsilon$\\
21 & $\lambda_{\min}(\cdot)$ & function that returns minimum eigenvalue of a matrix \\
22 & $\lambda$ & regularization parameter \\
23 & $\bA, \bb$ & sufficient statistics of $\cH$\\
24 & $\bM_{i,t}$ & user model of user $i$ at time $t$\\
25& $\tilde{k}_{i,t}$ & index of the unique bandit parameter associated with observations in $\bM_{i,t}$ \\
26 & $\bU_{t}, \bO_{t}$ & sets of up-to-date and outdated user models at time $t$ \\
27 & $\hat{V}_{i,t}$ & estimated neighborhood of user $i$ at time $t$ \\
28 & $\hat{N}^{\phi}_{\tilde{k}_{i,t}}(t)$ & the set of time steps associated with observations in $\hat{V}_{i,t}$\\
29 & $e_{i_{t},t}$ & indicator variable of the one-sample homogeneity test at time $t$\\
30 & $\hat{e}_{i,t}$ & empirical mean of $e_{i,t}$ in a sliding window, with its size denoted by $\tau$ \\
31 & $\hat{\theta}_{\hat{V}_{i,t}}$ & ridge regression estimator by aggregating observations in $\hat{V}_{i,t}$\\
32 & $CB_{\hat{V}_{i,t}}(\bx)$ & confidence bound for reward estimation on $\bx$ using $\hat{\theta}_{\hat{V}_{i,t}}$ \\
33 & $\upsilon^{e}, \upsilon^{c}$ & thresholds for homogeneity test in change detection and cluster identification \\
34 & $R_{Lin}(\cdot)$ & high probability regret upper bound of standard LinUCB \cite{abbasi2011improved}\\
\bottomrule
\end{tabular}
\label{tb:notations}
\end{table*}

\newpage

\section{Proof of Lemma \ref{lem:earlyDetection}}

Note that early detection corresponds to type-I error of the homogeneity test in Lemma \ref{lem:type1Similarity}, e.g., when change has not happened (thus $\cH_{i_{t},t-1}$ and $(\bx_{t},y_{t})$ are homogeneous), but the test statistic exceeds the threshold $\upsilon^{e}$: $e_{i_{t},t}=\textbf{1}\left\{s(\cH_{i_{t},t-1},\left\{(\bx_{t},y_{t})\right\})>\upsilon^{e}\right\}=1$. Therefore, we have $\bE[e_{i_{t},t}] \leq 1-F(\upsilon^{e};1,0)$. Then we can use Hoeffding inequality given in Lemma \ref{lem:hoeffding} to upper bound the early detection probability using $\hat{e}_{i_{t},t}$.

As the test variable $e_{i_{t},t} \in \left\{0,1\right\}$, it is $\frac{1}{2}$-sub-Gaussian. By applying Hoeffding inequality, we have:
\begin{equation*}
    P\big(\tau \hat{e}_{i_{t},t} - \tau \bE[e_{i_{t},t}] \geq h\big) \leq \exp{\Big(-\frac{2h^{2}}{\tau}\Big)}
\end{equation*}
Then setting $\delta_{e}=\exp{(-\frac{2h^{2}}{\tau})}$ gives $h=\sqrt{\frac{\tau \log{1/\delta_{e}}}{2}}$. Substituting this back and re-arrange the inequality gives us:
\begin{equation*}
P\Big(\hat{e}_{i_{t},t} < \bE[e_{i_{t},t}] + \sqrt{\frac{\log{1/\delta_{e}}}{2\tau }}\Big) > 1-\delta_{e}
\end{equation*}
Since $\bE[e_{i_{t},t}] \leq 1-F(\upsilon^{e};1,0)$, we have:
\begin{equation*}
P\Big(\hat{e}_{i_{t},t} < 1-F(\upsilon^{e};1,0) + \sqrt{\frac{\log{1/\delta_{e}}}{2\tau }}\Big) > 1-\delta_{e} \\
\end{equation*}
when change has not happened.

\section{Proof of Theorem \ref{thm:regretBound}}
In this section, we give the full proof of the upper regret bound in Theorem \ref{thm:regretBound}. 
We first define some additional notations necessary for the analysis and arrange the proof into three parts: 1) proof of Eq \eqref{eq_regret}; 2) proof of Lemma \ref{lem:secondTerm}; and 3) proof of Lemma \ref{lem:thirdTerm}. Specifically, Eq \eqref{eq_regret} provides an intermediate upper regret bound with three terms, and Lemma \ref{lem:secondTerm} and Lemma \ref{lem:thirdTerm} further bound the second and third terms to obtain the final result in Theorem \ref{thm:regretBound}.

Consider a learner that has access to the ground-truth change points and clustering structure, or equivalently, the learner knows the index of the unique bandit parameter each observation is associated with (but it does not know the value of the parameter). For example, when serving user $i_{t}$ at some time step $t$, the index of user $i_{t}$'s unique bandit parameter for the moment is $k_{i_{t},t}$, such that $\theta_{i_{t},t}=\phi_{k_{i_{t},t}}$. Then since this learner knows $k_{i_{t^{'}},t^{'}}$ for $t^{'} \in [t]$, it can precisely group the observations associated with each unique bandit parameter $\phi_{k}$ for $k\in[m]$. 
Recall that we denote $N^{\phi}_{k}(t)=\left\{1\leq t^{'} \leq t: \theta_{i_{t^{'}},t^{'}}=\phi_{k}\right\}$ as the set of time steps up to time $t$ when the user being served has the bandit parameter equal to $\phi_{k}$, e.g., all the observations obtained at time steps in $N^{\phi}_{k}(t)$ have the same unique bandit parameter $\phi_{k}$. Then an ideal reference algorithm would be the one that aggregates these observations to compute UCB score for arm selection. The regret of this ideal reference algorithm can be upper bounded by $\sum_{k=1}^{m}R_{Lin}(|N^{\phi}_{k}(T)|)$ where $R_{Lin}(|N^{\phi}_{k}(T)|)=O\Big(d\sqrt{|N^{\phi}_{k}(T)|\log^{2}{|N^{\phi}_{k}(T)|}}\Big)$  \cite{abbasi2011improved}.


However in our learning environment, such knowledge is not available to the learner; as a result, the learner does not know $N^{\phi}_{k_{i_{t},t}}(t-1)$ when interacting with user $i_{t}$ at time $t$; instead, it uses observations in the estimated neighborhood $\hat{V}_{i_{t},t-1}$ as shown in Algorithm \ref{algorithm} (line 17). Denote the set of time steps associated with observations in $\hat{V}_{i_{t},t-1}$ as $\hat{N}^{\phi}_{\tilde{k}_{i_{t},t-1}}(t-1)$, where $\tilde{k}_{i_{t},t-1}$ is the index of the unique parameter associated with observations in $\cH_{i_{t},t-1}$. We define a `good event' as $\left\{\hat{N}^{\phi}_{\tilde{k}_{i_{t},t-1}}(t-1)=N^{\phi}_{k_{i_{t},t}}(t-1)\right\}$, which matches with the reference algorithm, and since there is a non-zero probability of errors in both change detection and cluster identification, we also have `bad event' $\left\{\hat{N}^{\phi}_{\tilde{k}_{i_{t},t-1}}(t-1) \neq N^{\phi}_{k_{i_{t},t}}(t-1)\right\}$, which incurs extra regret.

Recall the estimated neighborhood $\hat{V}_{i_{t},t-1}=\left\{\bM \in \bU_{t-1}\cup\bO_{t-1}:S(\cH_{i_{t},t-1},\cH) \leq \upsilon^{c}\right\}$. If $k_{i_{t},t} \neq \tilde{k}_{i_{t},t-1}$, it means there is a mismatch between user model $\bM_{i_{t},t-1}$ and the current ground-truth user parameter $\theta_{i_{t},t}$, but the change detection module has failed to detect this. Then the obtained neighborhood is incorrect even if the cluster identification model made no mistake. Therefore, the bad event can be further decomposed into 
\footnotesize
$\left(\left\{\tilde{k}_{i_{t},t-1} \neq k_{i_{t},t}\right\} \cap \left\{\hat{N}^{\phi}_{\tilde{k}_{i_{t},t-1}}(t-1) \neq N^{\phi}_{k_{i_{t},t}}(t-1)\right\}\right) \cup \left(\left\{\tilde{k}_{i_{t},t-1} = k_{i_{t},t}\right\} \cap \left\{\hat{N}^{\phi}_{\tilde{k}_{i_{t},t-1}}(t-1) \neq N^{\phi}_{k_{i_{t},t}}(t-1)\right\}\right)$. 
\normalsize
The first part is a subset of event $\left\{\tilde{k}_{i_{t},t-1} \neq k_{i_{t},t}\right\}$, which suggests late detection happens at time $t$. The second part indicates incorrectly estimated clustering for user $i_t$ at time $t$.

\paragraph{Discussions} Before we move on, we would like to provide some explanations about the use of $\left\{\tilde{k}_{i_{t},t-1} \neq k_{i_{t},t}\right\}$ to denote the event that the user's underlying bandit parameter has changed, but the learner failed to detect it, i.e., late detection. Recall that $\tilde{k}_{i_{t},t-1}$ is the index of the unique bandit parameter associated with observations in $\bM_{i_{t},t-1}$, while $k_{i_{t},t}$ is the index of the unique parameter that governs observation $(\bx_{t},y_{t})$ from user $i_{t}$ at time $t$. Our change detection mechanism in Algorithm \ref{algorithm} (line 9) is expected to replace model $\bM_{i_{t},t-1}$ if change has happened, thus ensuring $\left\{\tilde{k}_{i_{t},t} = k_{i_{t},t}\right\}$. However, when it fails to detect the change, it will cause $\left\{\tilde{k}_{i_{t},t} = k_{i_{t},t+1}\right\}$ in the next time step $t+1$, which means the user model $\bM_{i_{t},t-1}$ cannot reflect the behaviors or preferences of user $i_{t}$ at time $t$.

With detailed proof in Section \ref{subsec:proofEq3}, following the decomposition discussed above, we can obtain:
\begin{align}
\label{eq_regret}
 R_{T} \leq &O\Big(\sigma d\sum_{k \in [m]}\sqrt{|N^{\phi}_{k}(T)|\log^{2}({|N^{\phi}_{k}(T)|})}\Big)+ 2\sum_{i \in \cU}\sum_{t \in \cN_{i}(T)} \textbf{1}\left\{\tilde{k}_{i_{t},t-1} \neq k_{i_{t},t}\right\} \\
\nonumber
& + 2\sum_{t=1}^{T}\textbf{1}\left\{\tilde{k}_{i_{t},t-1} = k_{i_{t},t}\right\} \cap \left\{\hat{N}^{\phi}_{\tilde{k}_{i_{t},t-1}}(t-1) \neq N^{\phi}_{k_{i_{t},t}}(t-1)\right\}
\end{align}
with a probability at least $1-\delta$.

In this upper regret bound, the first term matches the regret of the reference algorithm that has access to the exact change points and clustering structure of each user and time step. We can rewrite it using the frequency of unique model parameter $\phi_k$ as $O\big(\sigma d\sqrt{T\log{T}}(\sum_{k=1}^{m}\sqrt{p_{k}})\big)$ similar to Section A.4 in \cite{gentile2014online}.
The second term is the added regret caused by the late detection of change points; and the third term is the added regret caused by the incorrect cluster identification for arm selection. The latter two terms can be further bounded by the following lemmas. Their proofs are later given in Section \ref{subsec:proofSecondTerm} and Section \ref{subsec:proofThirdTerm} respectively.
\begin{lemma}\label{lem:secondTerm}
Under Assumption \ref{assump1} and \ref{assump3}, by setting the sliding window size \\
$\tau \geq \frac{2\log{1/\delta_{e}}}{\left\{[1-F(\upsilon^{e};1,\psi^{e})]\rho(1-\delta^{'})-1+F(\upsilon^{e};1,0)\right\}^{2}}$, where $\psi^{e}=\frac{\Delta^{2}/\sigma^{2}}{1+{1}/\Big[{\frac{\lambda^{'}}{4}S_{min}-8\big(\log{\frac{d S_{min}}{\delta^{'}}+\sqrt{S_{min}\log{\frac{d S_{min}}{\delta^{'}}}}}\big)}\Big]}$, the second term in Eq \eqref{eq_regret} can be upper bounded by:
\begin{equation*}
2\sum_{i \in \cU}\sum_{t \in \cN_{i}(T)} \textbf{1}\left\{\tilde{k}_{i_{t},t-1} \neq k_{i_{t}}\right\}
        \leq 2\sum_{i \in \cU}\Big(\Gamma_{i}(T)-1\Big)(\tau+\frac{2}{1-\delta_{e}})
\end{equation*}
with a probability at least $1-\frac{\delta_{e}}{1-\delta_{e}}$.
\end{lemma}

\begin{lemma}\label{lem:thirdTerm}
Define function $g(\psi;d,\upsilon)=F(\upsilon;\psi,d)$, and $g^{-1}(\cdot|d,\upsilon)$ as its inverse function.
Under Assumption \ref{assump2} and \ref{assump3}, the third term in Eq \eqref{eq_regret} can be upper bounded by: 
\begin{align*}
 2\sum_{t=1}^{T}\textbf{1}\left\{\tilde{k}_{i_{t},t-1} = k_{i_{t},t}\right\} \cap \left\{\hat{N}^{\phi}_{\tilde{k}_{i_{t},t-1}}(t-1) \neq N^{\phi}_{k_{i_{t},t}}(t-1)\right\}
 \leq 2 \sum_{i \in \cU} \Gamma_{i}(T)O\Big(\frac{2\psi^{c}\sigma^{2}}{\gamma^{2}{\lambda^{'}}^{2}}\log{\frac{d}{\delta^{'}}}\Big)
\end{align*}
with a probability at least $1-\delta^{'}$, where $\psi^{c}=g^{-1}\big(\frac{p(1-F(\upsilon^{c};d,0))}{1-p};d,\upsilon^{c}\big)$ is a constant.
\end{lemma}

Combining results in Eq \eqref{eq_regret}, Lemma \ref{lem:secondTerm} and Lemma \ref{lem:thirdTerm}, we obtain the upper regret bound:
\begin{align*}
\begin{split}
 R_{T} & \leq O\Big(\sigma d\sqrt{T\log^{2}{T}}(\sum_{k=1}^{m}\sqrt{p_{k}})\Big) + 2\sum_{i \in \cU}\Big(\Gamma_{i}(T)-1\Big)(\tau+\frac{2}{1-\delta_{e}}) + 2 \sum_{i \in \cU} \Gamma_{i}(T)O\Big(\frac{2\psi^{c}\sigma^{2}}{\gamma^{2}\lambda'^{2}}\log{\frac{d}{\delta^{'}}}\Big) \\
 & = O\Big(\sigma d\sqrt{T\log^{2}{T}}(\sum_{k=1}^{m}\sqrt{p_{k}})+\sum_{i \in \cU} \Gamma_{i}(T) \cdot C \Big)
\end{split}
\end{align*}
where $C=\frac{1}{1-\delta^{e}}+\frac{\sigma^{2}}{\gamma^{2}\lambda'^{2}}\log{\frac{d}{\delta^{'}}}$, with a probability at least $(1-\delta)(1-\frac{\delta_{e}}{1-\delta_{e}})(1-\delta^{'})$.

\subsection{Proof of Eq \eqref{eq_regret}}\label{subsec:proofEq3}

Recall that we define a `good' event as $\left\{\hat{N}^{\phi}_{\tilde{k}_{i_{t},t-1}}(t-1)=N^{\phi}_{k_{i_{t},t}}(t-1)\right\}$, which means at time $t$, \model{} selects an arm using the UCB score computed with all observations associated with $\phi_{k_{i_{t},t}}$. And the `bad' event is defined as its complement: $\left\{\hat{N}^{\phi}_{\tilde{k}_{i_{t},t-1}}(t-1) \neq N^{\phi}_{k_{i_{t},t}}(t-1)\right\}$, which can be decomposed and then contained as shown below:
\begin{equation*}
\begin{split}
& \left\{\hat{N}^{\phi}_{\tilde{k}_{i_{t},t-1}}(t-1) \neq N^{\phi}_{k_{i_{t},t}}(t-1)\right\} \\
 = &\left(\left\{\tilde{k}_{i_{t},t-1} \neq k_{i_{t},t}\right\} \cap \left\{\hat{N}^{\phi}_{\tilde{k}_{i_{t},t-1}}(t-1) \neq N^{\phi}_{k_{i_{t},t}}(t-1)\right\}\right) \\
& \cup \left(\left\{\tilde{k}_{i_{t},t-1} = k_{i_{t},t}\right\} \cap \left\{\hat{N}^{\phi}_{\tilde{k}_{i_{t},t-1}}(t-1) \neq N^{\phi}_{k_{i_{t},t}}(t-1)\right\}\right)\\
 \subseteq & \left\{\tilde{k}_{i_{t},t-1} \neq k_{i_{t},t}\right\} \cup \left(\left\{\tilde{k}_{i_{t},t-1} = k_{i_{t},t}\right\} \cap \left\{\hat{N}^{\phi}_{\tilde{k}_{i_{t},t-1}}(t-1) \neq N^{\phi}_{k_{i_{t},t}}(t-1)\right\}\right)\\
\end{split}
\end{equation*}

where the event $\bigl\{\tilde{k}_{i_{t},t-1} \neq k_{i_{t},t}\bigr\}$ means at time step $t$ there is a late detection, and the event $\bigl\{\tilde{k}_{i_{t},t-1} = k_{i_{t},t}\bigr\} \cap \bigl\{\hat{N}^{\phi}_{\tilde{k}_{i_{t},t-1}}(t-1) \neq N^{\phi}_{k_{i_{t},t}}(t-1)\bigr\}$ means there is no late detection, but the cluster identification fails to correctly cluster user models associated with $\phi_{k_{i_{t},t}}$ together (for example, there might be models not belonging to this cluster, or models failed to be put into this cluster).

Under the `good' event, arm $\bx_{t}$ is selected using the UCB strategy by aggregating all existing observations associated with $\phi_{k_{i_{t},t}}$, which is the unique bandit parameter for user $i_{t}$ at time $t$. To simplify the notations, borrowing the notation used in Eq \eqref{eq:UCB}, we denote $\hat{\theta}_{N^{\phi}_{k_{i_{t},t}}(t-1)}=\bA_{N^{\phi}_{k_{i_{t},t}}(t-1)}^{-1}\bb_{N^{\phi}_{k_{i_{t},t}}(t-1)}$, where $\bA_{N^{\phi}_{k_{i_{t},t}}(t-1)}=\lambda \bI + \sum_{j \in N^{\phi}_{k_{i_{t},t}}(t-1)}\bx_{j}\bx_{j}^{\top}$ and $\bb_{N^{\phi}_{k_{i_{t},t}}(t-1)}=\sum_{j \in N^{\phi}_{k_{i_{t},t}}(t-1)}\bx_{j}y_{j}$, as the ridge regression estimator, 
and $CB_{N^{\phi}_{k_{i_{t},t}}(t-1)}(\bx)=\alpha_{N^{\phi}_{k_{i_{t},t}}(t-1)}\sqrt{\bx^{\top} \bA_{N^{\phi}_{k_{i_{t},t}}(t-1)}^{-1}\bx}$, where $\alpha_{N^{\phi}_{k_{i_{t},t}}(t-1)}=\sigma\sqrt{d\log{(1+\frac{|N^{\phi}_{k_{i_{t},t}}(t-1)|}{d\lambda})} + 2\log{\frac{1}{\delta}}}+\sqrt{\lambda}$ is the corresponding reward estimation confidence bound on $\bx$.

Then we can upper bound the instantaneous regret $r_{t}$ as follows,
\begin{equation*}
\begin{split}
r_{t} & = \langle\theta_{i_{t},t},\bx_{t}^{*}\rangle - \langle\theta_{i_{t},t},\bx_{t}\rangle \\
&\leq \langle \tilde{\theta}_{i_{t},t},\bx_{t}\rangle - \langle\theta_{i_{t},t},\bx_{t}\rangle \\
& = \langle\tilde{\theta}_{i_{t},t}-\hat{\theta}_{\hat{V}_{i,t-1}},\bx_{t}\rangle + \langle\hat{\theta}_{\hat{V}_{i,t-1}}-\theta_{i_{t},t},\bx_{t}\rangle \\
& \leq \begin{cases}
    2 CB_{N^{\phi}_{k_{i_{t},t}}(t-1)}(\bx_{t}), & \text{if $\left\{\hat{N}^{\phi}_{\tilde{k}_{i_{t},t-1}}(t-1)=N^{\phi}_{k_{i_{t},t}}(t-1)\right\}$}.\\
    2, & \text{otherwise}.
  \end{cases}
\end{split}
\end{equation*}
The first inequality is because $\langle\tilde{\theta}_{i_{t},t},\bx_{t}\rangle$ is optimistic, where $\bx_{t} \in C_{t}$ and\\
$\tilde{\theta}_{i_{t},t} \in \left\{\theta \in \bR^{d}: \lVert \hat{\theta}_{\hat{V}_{i,t-1}}-\theta \rVert_{\bA_{\hat{V}_{i,t-1}}^{-1}} \leq \alpha_{N^{\phi}_{k_{i_{t},t}}(t-1)}\right\}$. For the second inequality, we split it into two cases according to the occurrence of the `good' or `bad' events. Recall that $\hat{N}^{\phi}_{\tilde{k}_{i_{t},t-1}}(t-1)$ denotes the set of time steps associated with observations in $\hat{V}_{i_{t},t-1}$. Then under the `good' event $\left\{\hat{N}^{\phi}_{\tilde{k}_{i_{t},t-1}}(t-1)=N^{\phi}_{k_{i_{t},t}}(t-1)\right\}$, with probability at least $1-\delta$, we have $\langle\tilde{\theta}_{i_{t},t}-\hat{\theta}_{\hat{V}_{i,t-1}},\bx_{t}\rangle \leq CB_{N^{\phi}_{k_{i_{t},t}}(t-1)}(\bx_{t})$ and $\langle\hat{\theta}_{\hat{V}_{i,t-1}}-\theta_{i_{t},t},\bx_{t}\rangle \leq CB_{N^{\phi}_{k_{i_{t},t}}(t-1)}(\bx_{t})$, so that $r_{t} \leq 2CB_{N^{\phi}_{k_{i_{t},t}}(t-1)}(\bx_{t})$. Under the `bad' event when wrong cluster is used for arm selection, we simply bound $r_{t}$ by 2 because $\lVert \theta_{i_{t},t} \rVert \leq 1$ and $\lVert \bx_{t} \rVert \leq 1$.

Then the accumulated regret $R_{T}$ can be upper bounded by:
\begin{equation*}\label{eq:cumulativeDecompose}
\begin{split}
R_{T} =& \sum_{t=1}^{T}r_{t} \\
\leq& 2 \sum_{t=1}^{T} \textbf{1}\left\{\hat{N}^{\phi}_{\tilde{k}_{i_{t},t-1}}(t-1)=N^{\phi}_{k_{i_{t},t}}(t-1)\right\} CB_{N^{\phi}_{k_{i_{t},t}}(t-1)}(\bx_{t}) \\
& + 2\sum_{t=1}^{T} \textbf{1}\left\{\hat{N}^{\phi}_{\tilde{k}_{i_{t},t-1}}(t-1) \neq N^{\phi}_{k_{i_{t},t}}(t-1)\right\} \\
\leq & \sum_{t=1}^{T} 2 CB_{N^{\phi}_{k_{i_{t},t}}(t-1)}(\bx_{t}) + 2\sum_{t=1}^{T} \textbf{1}\left\{\hat{N}^{\phi}_{\tilde{k}_{i_{t},t-1}}(t-1) \neq N^{\phi}_{k_{i_{t},t}}(t-1)\right\} \\
\leq & \sum_{t=1}^{T} 2 CB_{N^{\phi}_{k_{i_{t},t}}(t-1)}(\bx_{t}) + 2\sum_{t=1}^{T} \textbf{1}\left\{\tilde{k}_{i_{t},t-1} \neq k_{i_{t},t}\right\} \\
& + 2\sum_{t=1}^{T}\left(\textbf{1}\left\{\tilde{k}_{i_{t},t-1} = k_{i_{t},t}\right\} \cap \left\{\hat{N}^{\phi}_{\tilde{k}_{i_{t},t-1}}(t-1) \neq N^{\phi}_{k_{i_{t},t}}(t-1)\right\}\right)\\
\leq & \sum_{t=1}^{T} 2 CB_{N^{\phi}_{k_{i_{t},t}}(t-1)}(\bx_{t}) + 2\sum_{i \in \cU}\sum_{t \in \cN_{i}(T)} \textbf{1}\left\{\tilde{k}_{i,t-1} \neq k_{i,t}\right\} \\
& + 2\sum_{t=1}^{T}\left(\textbf{1}\left\{\tilde{k}_{i_{t},t-1} = k_{i_{t},t}\right\} \cap \left\{\hat{N}^{\phi}_{\tilde{k}_{i_{t},t-1}}(t-1) \neq N^{\phi}_{k_{i_{t},t}}(t-1)\right\}\right)
\end{split}
\end{equation*}
The first term is essentially the upper regret bound of the reference algorithm mentioned in Section \ref{subsec:regretAnalysis}, which can be further upper bounded with probability at least $1-\delta$ by:
\begin{equation*}
\begin{split}
& \sum_{t=1}^{T} 2 CB_{N^{\phi}_{k_{i_{t},t}}(t-1)}(\bx_{t}) = \sum_{k \in [m]}\sum_{t \in N^{\phi}_{k}(T)} 2 CB_{N^{\phi}_{k}(t-1)}(x_{t}) \leq \sum_{k \in [m]} R_{Lin}(|N^{\phi}_{k}(T)|) \\
\end{split}
\end{equation*}
where $R_{Lin}(|N^{\phi}_{k}(T)|)$ is the high probability upper regret bound in \cite{abbasi2011improved} (Theorem 3), which is defined as:
\begin{align*}
     R_{Lin}(|N^{\phi}_{k}(T)|) & =4\sqrt{d|N^{\phi}_{k}(T)|\log{(\lambda+\frac{|N^{\phi}_{k}(T)|}{d})}}\left(\sigma\sqrt{2\log{\frac{1}{\delta}}+d\log{(1+\frac{|N^{\phi}_{k}(T)|}{d\lambda})}} + \lambda^{1/2}\right) \\
    & = O\left(\sigma d \sqrt{|N^{\phi}_{k}(T)|\log^{2}{|N^{\phi}_{k}(T)|}} + \sigma \sqrt{d|N^{\phi}_{k}(T)|\log{\frac{|N^{\phi}_{k}(T)|}{\delta}}}\right)
\end{align*}

\subsection{Proof of Lemma \ref{lem:secondTerm}} \label{subsec:proofSecondTerm}
Now we have proved the intermediate regret upper bound in Eq \eqref{eq_regret}. In this section, we continue to upper bound its second term $2\sum_{i \in \cU}\sum_{t \in \cN_{i}(T)} \textbf{1}\left\{\tilde{k}_{i,t-1} \neq k_{i,t}\right\}$, which essentially counts the total number of late detections in each user, e.g., there is a mismatch between $\bM_{i_{t},t-1}$ and the current ground-truth bandit parameter $\theta_{i_{t},t}$, but the learner fails to detect this. To prove this lemma, we need the following lemmas that upper bound the probability of late detections.

As opposed to early detection in Lemma \ref{lem:earlyDetection}, late detection corresponds to type-II error of homogeneity test in Lemma \ref{lem:type2Similarity}. Therefore we have the following lemma.

\begin{lemma}\label{lem:lowerBoundProbDetection}
When change has happened ($\tilde{k}_{i_{t},t-1} \neq k_{i_{t},t}$), we have
\begin{equation*}
    P\big(e_{i_{t},t}=1\big) \geq \rho(1-\delta^{'})\big[1-F(\upsilon^{e};1,\psi^{e})\big]
\end{equation*}
where $\psi^{e}=\frac{\Delta^{2}/\sigma^{2}}{1+{1}/\big(\frac{\lambda^{'}}{4}S_{min}-8\big(\log{\frac{d S_{min}}{\delta^{'}}+\sqrt{S_{min}\log{\frac{d S_{min}}{\delta^{'}}}}}\big)\big)}$.
\end{lemma}

\textit{Proof of Lemma \ref{lem:lowerBoundProbDetection}.}

Combining Lemma \ref{lem:type2Similarity}, Assumption \ref{assump1} and \ref{assump3}, we can lower bound the probability that $e_{i_{t},t}=1$ when change has happened as:
\begin{equation*}
\begin{split}
 P\big(e_{i_{t},t}=1\big) & =P\big(s(\cH_{i_{t},t-1},\left\{\bx_{t},y_{t}\right\})>\upsilon^{e}\big) \\
& \geq 1-F\left(\upsilon^{e};1,\frac{[\bx_{t}^{\top}(\theta_{i_{t},c}-\theta_{i_{t},c-1})]^{2}/\sigma^{2}}{1+\bx_{t}^{\top}(\sum_{(\bx_{k},y_{k}) \in \cH_{i_{t},t-1}}\bx_{k}\bx_{k}^{\top})^{-1}\bx_{t}}\right) \\
& \geq 1-F\left(\upsilon^{e};1,\frac{[\bx_{t}^{\top}(\theta_{i_{t},c}-\theta_{i_{t},c-1})]^{2}/\sigma^{2}}{1+\frac{||\bx_{t}||^{2}}{\lambda_{\min}(\sum_{(\bx_{k},y_{k}) \in \cH_{i_{t},t-1}}\bx_{k}\bx_{k}^{\top})}}\right) \\
& \geq \rho\left[1-F\left(\upsilon^{e};1,\frac{\Delta^{2}/\sigma^{2}}{1+\frac{1}{\lambda_{\min}(\sum_{(\bx_{k},y_{k}) \in \cH_{i_{t},t-1}}\bx_{k}\bx_{k}^{\top})}}\right)\right] \\
\end{split}
\end{equation*}
Since the minimum length of stationary period is $S_{min}$, by applying Lemma \ref{lem:minimumEigenvalueLB}, we can obtain the following lower bound on minimum eigenvalue when change happens as:
\begin{equation*}
    \lambda_{\min}\left(\sum_{(\bx_{k},y_{k}) \in \cH_{i_{t},t-1}}\bx_{k}\bx_{k}^{\top}\right) \geq \frac{\lambda^{'}}{4}S_{min}-8\left(\log{\frac{d S_{min}}{\delta^{'}}+\sqrt{S_{min}\log{\frac{d S_{min}}{\delta^{'}}}}}\right)
\end{equation*}
with probability at least $1-\delta^{'}$.

Denote $\psi^{e}=\frac{\Delta^{2}/\sigma^{2}}{1+{1}/\big(\frac{\lambda^{'}}{4}S_{min}-8\big(\log{\frac{d S_{min}}{\delta^{'}}+\sqrt{S_{min}\log{\frac{d S_{min}}{\delta^{'}}}}}\big)\big)}$. We obtain the following lower bound on the probability of detection:
\begin{equation*}
    P\big(e_{i_{t},t}=1\big) \geq \rho(1-\delta^{'})\big[1-F(\upsilon^{e};1,\psi^{e})\big]
\end{equation*}
when change has happened.

\begin{lemma}\label{lem:lateDetection}
When change has happened ($\tilde{k}_{i_{t},t-1} \neq k_{i_{t},t}$),
\begin{equation*}
    P\left(\hat{e}_{i_{t},t} \geq 1-F(\upsilon^{e};1,0) + \sqrt{\frac{\log{1/\delta_{e}}}{2\tau }}\right) \geq 1-\delta_{e}
\end{equation*}
if the size of sliding window $\tau \geq \frac{2\log{1/\delta_{e}}}{\left\{[1-F(\upsilon^{e};1,\psi^{e})]\rho(1-\delta^{'})-1+F(\upsilon^{e};1,0)\right\}^{2}}$.
\end{lemma}

\textit{Proof of Lemma \ref{lem:lateDetection}.}

Similarly to the proof of Lemma \ref{lem:earlyDetection}, applying Hoeffding inequality given in Lemma \ref{lem:hoeffding}, we have:
\begin{equation*}
\begin{split}
    P\left(\hat{e}_{i_{t},t}  \leq \bE[e_{i,t}] - \sqrt{\frac{\log{1/\delta_{e}}}{2\tau }}\right) &\leq \delta_{e} \\
    P\left(\hat{e}_{i_{t},t}  \geq \bE[e_{i,t}] - \sqrt{\frac{\log{1/\delta_{e}}}{2\tau}}\right) &\geq 1-\delta_{e} \\
\end{split}
\end{equation*}
From Lemma \ref{lem:lowerBoundProbDetection}, when change has happened, $\bE[e_{i_{t},t}] \geq \rho(1-\delta^{'})\left[1-F(\upsilon^{e};1,\psi^{e})\right]$, with $\psi^{e}$ being a variable dependent on environment as defined in Lemma \ref{lem:lowerBoundProbDetection}. By substituting this into the above inequality, we have:
\begin{equation*}
    P\left(\hat{e}_{i,t} \geq \rho(1-\delta^{'})\left[1-F(\upsilon^{e};1,\psi^{e})\right] - \sqrt{\frac{\log{1/\delta_{e}}}{2\tau }}\right) \geq 1-\delta_{e}
\end{equation*}
Then by rearranging terms above, we can find that if the sliding window size $\tau$ is selected to satisfy:
\begin{equation*}
    \tau \geq \frac{2\log{1/\delta_{e}}}{\Big\{[1-F(\upsilon^{e};1,\psi^{e})]\rho(1-\delta^{'})-1+F(\upsilon^{e};1,0)\Big\}^{2}}
\end{equation*}
we can obtain:
\begin{equation*}
\begin{split}
P\left(\hat{e}_{i_{t},t} \geq 1-F(\upsilon^{e};1,0) + \sqrt{\frac{\log{1/\delta_{e}}}{2\tau }}\right) & \geq 1-\delta_{e} \\
P\left(\hat{e}_{i_{t},t} \leq 1-F(\upsilon^{e};1,0) + \sqrt{\frac{\log{1/\delta_{e}}}{2\tau }}\right) & \leq \delta_{e} \\
\end{split}
\end{equation*}
when change has happened ($\tilde{k}_{i_{t},t-1} \neq k_{i_{t},t}$).

\textit{Proof of Lemma \ref{lem:secondTerm}.}

With results from Lemma \ref{lem:lateDetection}, our solution to further upper bound the number of late detections in each stationary period is similar to \cite{wu2018learning} (Theorem 3.2). We include the proof here for the sake of completeness. 


Denote the probability of detection when change has happened as $p_{d}=P\Big(\hat{e}_{i_{t},t} \geq 1-F(\upsilon^{e};1,0) + \sqrt{\frac{\log{1/\delta_{e}}}{2\tau }}\Big)$, and from Lemma \ref{lem:lateDetection}, we have $p_{d} \geq 1-\delta_{e}$. The probability distribution over the number of late detections when change has happened follows a geometric distribution: $P(n_\text{late}=k)=(1-p_{d})^{k-1}p_{d}$. Then by applying Chebyshev's inequality, we have $P\Big(n_\text{late} \leq \frac{2}{1-\delta_{e}}\Big) \geq 1-\frac{\delta_{e}}{1-\delta_{e}}$.

Now we can upper bound the number of late detections $\sum_{i \in \cU}\sum_{t \in \cN_{i}(T)} \textbf{1}\left\{\tilde{k}_{i_{t},t-1} \neq k_{i_{t},t}\right\}$ in user $i$. In Assumption
\ref{assump1} we have assumed that the total number of change points of user $i$ is $\Gamma_{i}(T)-1$. Therefore, $\sum_{t \in \cN_{i}(T)} \textbf{1}\big\{\tilde{k}_{i_{t},t-1} \neq k_{i_{t}}\big\} \leq \big(\Gamma_{i}(T)-1 \big)(\tau+\frac{2}{1-\delta_{e}})$ with probability at least $1-\frac{\delta_{e}}{1-\delta_{e}}$. Then we can upper bound the second term in Eq \eqref{eq_regret} by:
\begin{equation*}
\begin{split}
        & 2\sum_{i \in \cU}\sum_{t \in \cN_{i}(T)} \textbf{1}\left\{\tilde{k}_{i_{t},t-1} \neq k_{i_{t},t}\right\} \leq 2\sum_{i \in \cU}\big(\Gamma_{i}(T)-1\big)(\tau+\frac{2}{1-\delta_{e}})
\end{split}
\end{equation*}

\subsection{Proof of Lemma \ref{lem:thirdTerm}}\label{subsec:proofThirdTerm}
The third term $\sum_{t=1}^{T}\textbf{1}\left\{\tilde{k}_{i_{t},t-1} = k_{i_{t},t}\right\} \cap \left\{\hat{N}^{\phi}_{\tilde{k}_{i_{t},t-1}}(t-1) \neq N^{\phi}_{k_{i_{t},t}}(t-1)\right\}$ counts the total number of times that there is no late detection, but cluster identification module fails to correctly cluster user models. We upper bound this using a similar technique as \cite{gentile2017context}. For the proof of Lemma \ref{lem:thirdTerm}, we need the following lemmas related to probability of errors of cluster identification.
\begin{lemma}\label{lem:type1Cluster}
When the underlying bandit parameters $\phi_{\tilde{k}_{i,t-1}}$ and $\phi_{\tilde{k}_{j,t-1}}$ of two observation history $\cH_{i,t-1}$ and $\cH_{j,t-1}$ are the same, the probability that cluster identification fails to cluster them together corresponds to the type-I error probability given in Lemma \ref{lem:type1Similarity}, and it can be upper bounded by:
\begin{equation*}
P\Big(S(\cH_{i,t-1},\cH_{j,t-1}) > \upsilon^{c}\mid\phi_{\tilde{k}_{i,t-1}}=\phi_{\tilde{k}_{j,t-1}}\Big) \leq 1-F(\upsilon^{c};df,0)
\end{equation*}
where $df=rank(\bX_{1})+rank(X_{2})-rank(\left[\begin{matrix}X_{1} \\ X_{2}\end{matrix}\right])$.
\end{lemma}
\begin{corollary}[Lower bound $P\big(N^{\phi}_{k_{i_{t},t}}(t-1) \subseteq \hat{N}^{\phi}_{\tilde{k}_{i_{t},t-1}}(t-1)\big)$]\label{lem:c1}
Since $N^{\phi}_{k_{i_{t},t}}(t-1)$ denotes the set of time indices associated with all observations whose underlying bandit parameter is $\phi_{k_{i_{t},t}}$, and $\hat{N}^{\phi}_{\tilde{k}_{i,t-1}}(t-1)$ denotes 
those in the estimated neighborhood $\hat{V}_{i_{t},t-1}$, when there is no late detection, i.e., we have $\tilde{k}_{i_{t},t-1} = k_{i_{t},t}$.
It naturally follows Lemma \ref{lem:type1Cluster} that:
\begin{equation*}
    P\big(N^{\phi}_{k_{i_{t}}}(t-1) \subseteq \hat{N}^{\phi}_{\tilde{k}_{i_{t},t-1}}(t-1)\big) \geq F(\upsilon^{c};df,0)
\end{equation*}
\end{corollary}
\begin{lemma}\label{lem:type2Cluster}
When the underlying bandit parameters $\phi_{\tilde{k}_{i,t-1}}$ and $\phi_{\tilde{k}_{j,t-1}}$ of two observation sequence $\cH_{i,t-1}$ and $\cH_{j,t-1}$ are not the same, the probability that cluster identification module clusters them together corresponds to the type-II error probability given in Lemma \ref{lem:type2Similarity}, which can be upper bounded by:
\begin{equation*}
P\Big(S(\cH_{i,t-1},\cH_{j,t-1}) \leq \upsilon^{c}|\phi_{\tilde{k}_{i,t-1}} \neq \phi_{\tilde{k}_{j,t-1}}\Big) \leq F(\upsilon^{c};d,\psi^{c})
\end{equation*}
under the condition that both $\lambda_{\min}(\sum_{(\bx_{k},y_{k}) \in \cH_{i,t-1}}\bx_{k}\bx_{k}^{\top})$ and $\lambda_{\min}(\sum_{(\bx_{k},y_{k}) \in \cH_{j,t-1}}\bx_{k}\bx_{k}^{\top})$ are at least $\frac{2\psi^{c}\sigma^{2}}{\gamma^{2}}$.
\end{lemma}

\textit{Proof of Lemma \ref{lem:type2Cluster}.}

Recall that type-II error probability of the homogeneity test can be upper bounded by $P\big(S(\cH_{i,t-1}, \\ \cH_{j,t-1}) \leq \upsilon^{c}|\phi_{\tilde{k}_{i,t-1}} \neq \phi_{\tilde{k}_{j,t-1}}\big) \leq F(\upsilon^{c};df,\psi)$ as discussed in Section \ref{lem:type2Similarity}. If either design matrix of the two datasets is rank-deficient, the noncentrality parameter $\psi$ is lower bounded by $0$ (lower bound achieved when the difference between two parameters lies in the null space of rank-deficient design matrix). 
Therefore, a nontrivial upper bound of type-II error probability only exists when the design matrices of both datasets are rank-sufficient. In this case, combining Lemma \ref{lem:type2Similarity} and Assumption \ref{assump2} gives:
\begin{align*}
 & P\big(S(\cH_{i,t-1},\cH_{j,t-1})\big) \\
 \leq & F\left(\upsilon^{c};d,\frac{||\phi_{\tilde{k}_{i,t-1}} - \phi_{\tilde{k}_{j,t-1}}||^{2}/\sigma^{2}}{{1}/{\lambda_{\min}(\sum_{(\bx_{k},y_{k}) \in \cH_{i,t-1}}\bx_{k}\bx_{k}^{\top})}+{1}/{\lambda_{\min}(\sum_{(\bx_{k},y_{k}) \in \cH_{j,t-1}}\bx_{k}\bx_{k}^{\top})}}\right) \\
    \leq & F\left(\upsilon^{c};d,\frac{\gamma^{2}/\sigma^{2}}{{1}/{\lambda_{\min}(\sum_{(\bx_{k},y_{k}) \in \cH_{i,t-1}}\bx_{k}\bx_{k}^{\top})}+{1}/{\lambda_{\min}(\sum_{(\bx_{k},y_{k}) \in \cH_{j,t-1}}\bx_{k}\bx_{k}^{\top})}}\right)
\end{align*}
Define $\psi^{c}>0$; then by rearranging terms we obtain the conditions that, when both:
\begin{equation*}
    \lambda_{\min}\left(\sum_{(\bx_{k},y_{k}) \in \cH_{i,t-1}}\bx_{k}\bx_{k}^{\top}\right) \geq \frac{2\psi^{c}\sigma^{2}}{\gamma^{2}} \text{~~~and~~~}
    \lambda_{\min}\left(\sum_{(\bx_{k},y_{k}) \in \cH_{j,t-1}}\bx_{k}\bx_{k}^{\top}\right) \geq \frac{2\psi^{c}\sigma^{2}}{\gamma^{2}} 
\end{equation*}
we have $$P\big(S(\cH_{i,t-1},\cH_{j,t-1}) \leq \upsilon^{c}|\phi_{\tilde{k}_{i,t-1}} \neq \phi_{\tilde{k}_{j,t-1}}\big) \leq F(\upsilon^{c};d,\psi^{c})$$

\begin{lemma}\label{lem:RealNegativeOutOfPredictedNegative}
If the cluster identification module clusters observation history $\cH_{i,t-1}$ and $\cH_{j,t-1}$ together, the probability that they actually have the same underlying bandit parameters is denoted as $P\big(\phi_{\tilde{k}_{i,t-1}}=\phi_{\tilde{k}_{j,t-1}}|S(\cH_{i,t-1},\cH_{j,t-1}) \leq \upsilon^{c}\big)$.
\begin{equation*}
    P\big(\phi_{\tilde{k}_{i,t-1}}=\phi_{\tilde{k}_{j,t-1}}|S(\cH_{i,t-1},\cH_{j,t-1}) \leq \upsilon^{c}\big) \geq F(\upsilon^{c};df,0)
\end{equation*}
under the condition that both $\lambda_{\min}\big(\sum_{(\bx_{k},y_{k}) \in \cH_{i,t-1}}\bx_{k}\bx_{k}^{\top}\big)$ and $\lambda_{\min}\big(\sum_{(\bx_{k},y_{k}) \in \cH_{j,t-1}}\bx_{k}\bx_{k}^{\top}\big)$ are at least $\frac{2\psi^{c}\sigma^{2}}{\gamma^{2}}$, where $\psi^{c}=g^{-1}\big(\frac{p(1-F(\upsilon^{c};d,0))}{1-p};d,\upsilon^{c}\big)$.
\end{lemma}

\textit{Proof of Lemma \ref{lem:RealNegativeOutOfPredictedNegative}.}

\paragraph{Discussions} Compared with the type-I and type-II error probabilities given in Lemma \ref{lem:type1Cluster} and \ref{lem:type2Cluster}, the probability $P(\phi_{\tilde{k}_{i,t-1}}=\phi_{\tilde{k}_{j,t-1}}|S(\cH_{i,t-1},\cH_{j,t-1}) \leq \upsilon^{c})$ also depends on the population being tested on. Two extreme examples would be 1) testing on a population that all user models have the same bandit parameter, and 2) every user model has an unique bandit parameter. Then in the former case $P(\phi_{\tilde{k}_{i,t-1}}=\phi_{\tilde{k}_{j,t-1}}|S(\cH_{i,t-1},\cH_{j,t-1}) \leq \upsilon^{c})=1$ and in the latter case $P(\phi_{\tilde{k}_{i,t-1}}=\phi_{\tilde{k}_{j,t-1}}|S(\cH_{i,t-1},\cH_{j,t-1}) \leq \upsilon^{c})=0$.

Denote the events $\big\{\phi_{\tilde{k}_{i,t-1}} \neq \phi_{\tilde{k}_{j,t-1}}\big\} \cap \big\{S(\cH_{i,t-1},\cH_{j,t-1}) > \upsilon^{c}\big\}$ as True Positive ($TP$), $\big\{\phi_{\tilde{k}_{i,t-1}} = \phi_{\tilde{k}_{j,t-1}}\big\} \cap \big\{S(\cH_{i,t-1},\cH_{j,t-1}) \leq \upsilon^{c}\big\}$ as True Negative ($TN$), $\big\{\phi_{\tilde{k}_{i,t-1}} = \phi_{\tilde{k}_{j,t-1}}\big\} \cap \big\{S(\cH_{i,t-1},\cH_{j,t-1}) > \upsilon^{c}\big\}$ as False Positive ($FP$), and $\big\{\phi_{\tilde{k}_{i,t-1}} \neq \phi_{\tilde{k}_{j,t-1}}\big\} \cap \big\{S(\cH_{i,t-1},\cH_{j,t-1}) \leq \upsilon^{c}\big\}$ as False Negative ($FN$) of cluster identification, respectively. We can rewrite the probabilities in Lemma \ref{lem:type1Cluster}, \ref{lem:type2Cluster} and \ref{lem:RealNegativeOutOfPredictedNegative} as:
\begin{align*}
P\big(S(\cH_{i,t-1},\cH_{j,t-1}) > \upsilon^{c}|\phi_{\tilde{k}_{i,t-1}}=\phi_{\tilde{k}_{j,t-1}}\big) &= \frac{FP}{TN+FP} \leq 1-F(\upsilon^{c};df,0) \\
P\big(S(\cH_{i,t-1},\cH_{j,t-1}) \leq \upsilon^{c}|\phi_{\tilde{k}_{i,t-1}} \neq \phi_{\tilde{k}_{j,t-1}}\big) &= \frac{FN}{FN+TP} \leq F(\upsilon^{c};df,\psi^{c}) \\
P\big(\phi_{\tilde{k}_{i,t-1}}=\phi_{\tilde{k}_{j,t-1}}|S(\cH_{i,t-1},\cH_{j,t-1}) \leq \upsilon^{c}\big) &= \frac{TN}{TN+FN} = \frac{1}{1+\frac{FN}{TN}}
\end{align*}
We can upper bound $\frac{FN}{TN}$ by:
\begin{equation*}
\frac{FN}{TN} \leq \frac{TP+FN}{TN+FP} \cdot \frac{F(\upsilon^{c};df,\psi^{c})}{F(\upsilon^{c};df,0)}
\end{equation*}
where $\frac{TP+FN}{TN+FP}$ denotes the ratio between the number of positive instances ($\phi_{\tilde{k}_{i,t-1}} \neq \phi_{\tilde{k}_{j,t-1}}$) and negative instances ($\phi_{\tilde{k}_{i,t-1}}=\phi_{\tilde{k}_{j,t-1}}$) in the population, which can be upper bounded by $\frac{1-p}{p}$ where $p=\min_{k}p_{k}$.

It is worth noting that when either design matrix of $\cH_{i,t-1}$ or $\cH_{j,t-1}$ does not have full column rank, $P\big(\phi_{\tilde{k}_{i,t-1}}=\phi_{\tilde{k}_{j,t-1}}|S(\cH_{i,t-1},\cH_{j,t-1}) \leq \upsilon^{c}\big) \geq {1}/\big(1+\frac{1-p}{p}\cdot \frac{F(\upsilon^{c};df,0)}{F(\upsilon^{c};df,0)}\big) \geq p$, which is then trivially lower bounded by the percentage of negative instances in the population.

Under the conditions given in Lemma \ref{lem:type2Cluster}, we have:
\begin{equation*}
P\big(\phi_{\tilde{k}_{i,t-1}}=\phi_{\tilde{k}_{j,t-1}}|S(\cH_{i,t-1},\cH_{j,t-1}) \leq \upsilon^{c}\big) \geq {1}/\Big(1+ \frac{1-p}{p} \cdot \frac{F(\upsilon^{c};d,\psi^{c})}{F(\upsilon^{c};d,0)}\Big)
\end{equation*}
Then by setting $\psi^{c}=g^{-1}\big(\frac{p(1-F(\upsilon^{c};d,0))}{1-p};d,\upsilon^{c}\big)$, we have:
\begin{equation*}
    P\big(\phi_{\tilde{k}_{i,t-1}}=\phi_{\tilde{k}_{j,t-1}}|S(\cH_{i,t-1},\cH_{j,t-1}) \leq \upsilon^{c}\big) \geq {1}/\Big(1+ \frac{1-p}{p} \cdot \frac{F(\upsilon^{c};d,\psi^{c})}{F(\upsilon^{c};d,0)}\Big) = F(\upsilon^{c};df,0)
\end{equation*}
\begin{corollary}[Lower bound $P\big(\hat{N}^{\phi}_{\tilde{k}_{i_{t},t-1}}(t-1) \subseteq N^{\phi}_{k_{i_{t},t}}(t-1)\big)$]\label{lem:c2}
It naturally follows Lemma \ref{lem:RealNegativeOutOfPredictedNegative} that:
\begin{equation*}
    P\big(\hat{N}^{\phi}_{\tilde{k}_{i_{t},t-1}}(t-1) \subseteq N^{\phi}_{k_{i_{t},t}}(t-1)\big) \geq F(\upsilon^{c};df,0)
\end{equation*}
under the condition that both $\lambda_{\min}(\sum_{(\bx_{k},y_{k}) \in \cH_{i,t-1}}\bx_{k}\bx_{k}^{\top})$ and $\lambda_{\min}(\sum_{(\bx_{k},y_{k}) \in \cH_{j,t-1}}\bx_{k}\bx_{k}^{\top})$ are at least $\frac{2\psi^{c}\sigma^{2}}{\gamma^{2}}$, where $\psi^{c}=g^{-1}\big(\frac{p(1-F(\upsilon^{c};d,0))}{1-p};d,\upsilon^{c}\big)$.

\end{corollary}

\textit{Proof of Lemma \ref{lem:thirdTerm}.}

From Corollary \ref{lem:c1} and \ref{lem:c2}, when both $\lambda_{\min}(\sum_{(\bx_{k},y_{k}) \in \cH_{i,t-1}}\bx_{k}\bx_{k}^{\top})$ and $\lambda_{\min}(\sum_{(\bx_{k},y_{k}) \in \cH_{j,t-1}}\bx_{k}\bx_{k}^{\top})$ are at least $\frac{2\psi^{c}\sigma^{2}}{\gamma^{2}}$, with probability at least $F(\upsilon^{c};df,0)$, we have event 
$\bigl\{\tilde{k}_{i_{t},t-1} =  k_{i_{t},t}\bigr\} \cap \bigl\{\hat{N}^{\phi}_{\tilde{k}_{i_{t},t-1}}(t-1) =  N^{\phi}_{k_{i_{t},t}}(t-1)\bigr\}$. 
Therefore, the third term in Eq \eqref{eq_regret} is upper bounded by:
\begin{equation*}
\begin{split}
    & 2\sum_{t=1}^{T}\textbf{1}\left\{\tilde{k}_{i_{t},t-1} = k_{i_{t},t}\right\} \cap \left\{\hat{N}^{\phi}_{\tilde{k}_{i_{t},t-1}}(t-1) \neq N^{\phi}_{k_{i_{t},t}}(t-1)\right\} \\
     \leq & 2\sum_{t=1}^{T} \textbf{1}\left\{\exists \cH\in \bU_{t-1} \cup \bO_{t-1}: \lambda_{\min}\Big(\sum_{(\bx_{k},y_{k}) \in \cH}\bx_{k}\bx_{k}^\top\Big) < \frac{2\psi^{c}\sigma^{2}}{\gamma^{2}} \right\} \\
     \leq & 2\sum_{i \in \cU} \sum_{t \in \cN_{i}(T)} \textbf{1}\left\{\lambda_{\min}\Big(\sum_{(\bx_{k},y_{k}) \in \cH_{i,t-1}}\bx_{k}\bx_{k}^\top\Big) < \frac{2\psi^{c}\sigma^{2}}{\gamma^{2}} \right\}
\end{split}
\end{equation*}
Essentially, it counts the number of time steps in total when minimum eigenvalue of a user model $\bM$'s correlation matrix is smaller than $\frac{2\psi^{c}\sigma^{2}}{\gamma^{2}}$. We further decompose the summation by considering each stationary period of each user.
\begin{equation*}
\begin{split}
& 2\sum_{i \in \cU} \sum_{t \in \cN_{i}(T)} \textbf{1}\left\{\lambda_{\min}\Big(\sum_{(\bx_{k},y_{k}) \in \cH_{i,t-1}}\bx_{k}\bx_{k}^\top\Big) < \frac{2\psi^{c}\sigma^{2}}{\gamma^{2}} \right\} \\
 = & 2 \sum_{i \in \cU} \sum_{s \in [0, c_{i,1},..,c_{i,\Gamma_{i}(T)-1}]} \sum_{t \in S_{i,s}}\textbf{1}\left\{\lambda_{\min}\Big(\sum_{(\bx_{k},y_{k}) \in \cH_{i,t-1}}\bx_{k}\bx_{k}^\top\Big) < \frac{2\psi^{c}\sigma^{2}}{\gamma^{2}} \right\}
\end{split}
\end{equation*}
where $S_{i,s}$ denotes the $s$'th stationary period of user $i$.

Borrowing the notation from \cite{gentile2017context}, denote $A_{t}$ as a correlation matrix constructed through a series of rank-one updates using context vectors from $\left\{C_{t}\right\}_{t \in S}$, where $S$ denotes the set of time steps we performed model update. Note that the choice of which context vector to select from $C_{t}$ for $t\in S$ can be arbitrary. Then we denote the maximum number of updates it takes until $\lambda_{\min}(A_{t})$ is lower bounded by $\eta$ as $HD(\left\{C_{t}\right\}_{t \in S}, \eta) =\max\left\{t \in S: \exists \bx_{1} \in C_{1},...,\bx_{t} \in C_{t}: \lambda_{\min}(A_{t}) \leq \eta \right\}$, where $A_{t}=\sum_{u \in S: u \leq t}\bx_{u}\bx_{u}^{\top}$. Therefore, we obtain:
\begin{align*}
& \sum_{i \in \cU} \sum_{t \in \cN_{i}(T)} \textbf{1}\left\{\lambda_{\min}\Big(\sum_{(\bx_{k},y_{k}) \in \cH_{i,t-1}}\bx_{k}\bx_{k}^\top\Big) < \frac{2\psi^{c}\sigma^{2}}{\gamma^{2}} \right\} \\
 = & \sum_{i \in \cU} \sum_{s \in [0, c_{i,1},..,c_{i,\Gamma_{i}(T)-1}]} \sum_{t \in S_{i,s}}\textbf{1}\left\{\lambda_{\min}\Big(\sum_{(\bx_{k},y_{k}) \in \cH_{i,t-1}}\bx_{k}\bx_{k}^\top\Big) < \frac{2\psi^{c}\sigma^{2}}{\gamma^{2}} \right\} \\
\leq &  \sum_{i \in \cU} \sum_{s \in [0, c_{i,1},..,c_{i,\Gamma_{i}(T)-1}]} HD\Big(\left\{C_{t}\right\}_{t \in S_{i,s}},\frac{2\psi^{c}\sigma^{2}}{\gamma^{2}}\Big) \end{align*}

Then similar to \cite{gentile2017context} (Lemma 1), by applying Lemma \ref{lem:minimumEigenvalueLB} we can upper bound the third term in Eq \eqref{eq_regret}:
\begin{align*}
& 2\sum_{t=1}^{T}\textbf{1}\left\{\tilde{k}_{i_{t},t-1} = k_{i_{t},t}\right\} \cap \left\{\hat{N}^{\phi}_{\tilde{k}_{i_{t},t-1}}(t-1) \neq N^{\phi}_{k_{i_{t},t}}(t-1)\right\}\\
 \leq & 2 \sum_{i \in \cU} \sum_{s \in [0, c_{i,1},..,c_{i,\Gamma_{i}(T)-1}]} HD\Big(\left\{C_{t}\right\}_{t \in S_{i,s}},\frac{2\psi^{c}\sigma^{2}}{\gamma^{2}}\Big) \\
 \leq & 2 \sum_{i \in \cU} \Gamma_{i}(T)O\Big(\frac{2\psi^{c}\sigma^{2}}{\gamma^{2}{\lambda^{'}}^{2}}\log{\frac{d}{\delta^{'}}}\Big) 
\end{align*}
with probability at least $1-\delta^{'}$.

\section{Technical lemmas}
Here are some of the technical lemmas needed for the proofs in this paper.

\begin{lemma}[Hoeffding inequality]\label{lem:hoeffding}
Suppose that we have independent variables $x_{i},i=1,\dots,n$, and $x_{i}$ has mean $\mu_{i}$ and sub-Gaussian parameter $\sigma_{i}$. Then for all $h\geq 0$, we have
\begin{equation*}
    P\big(\sum_{i=1}^{n}(x_{i}-\mu_{i})\geq h\big) \leq \exp\left(-\frac{h^{2}}{2\sum_{i=1}^{n}\sigma_{i}^{2}}\right)
\end{equation*}
\end{lemma}

\begin{lemma}[Lemma 1 of \cite{gentile2017context}]\label{lem:minimumEigenvalueLB}
Under Assumption \ref{assump3} that, at each time $t$, arm set $C_{t}$ is generated i.i.d. from a sub-Gaussian random vector $X \in \bR^{d}$, such that $\bE[XX^{\top}]$ is full-rank with minimum eigenvalue $\lambda'>0$; and the variance $\varsigma^{2}$ of the random vector satisfies $\varsigma^{2} \leq \frac{{\lambda'}^{2}}{8\ln{4K}}$. Then we have the following lower bound on minimum eigenvalue of the correlation matrix of observation history $\cH$:
\begin{equation*}
    \lambda_{\min}\Big(\sum_{(\bx_{k},y_{k}) \in \cH}\bx_{k}\bx_{k}^{\top}\Big) \geq \frac{\lambda^{'}}{4}|\cH|-8\Big(\log{\frac{d |\cH|}{\delta^{'}}+\sqrt{|\cH|\log{\frac{d |\cH|}{\delta^{'}}}}}\Big)
\end{equation*}
with probability at least $1-\delta^{'}$.
\end{lemma}

\end{document}